%% file: neurips_2021.tex
\documentclass{article}

    \usepackage[preprint]{neurips_2021}

\usepackage[utf8]{inputenc} 
\usepackage[T1]{fontenc}    
\input{packages}

\input{macros}

\input{math_commands}

\usepackage{hyperref}
\usepackage{url}

\usepackage[capitalize,noabbrev]{cleveref}

\title{Pre-Trained Language Models for \\ Interactive Decision-Making}

\author{
Shuang Li \textsuperscript{1}\thanks{Correspondence to:
Shuang Li <lishuang@mit.edu>} \;, 
Xavier Puig\textsuperscript{1}, 
Chris Paxton\textsuperscript{2}, 
Yilun Du\textsuperscript{1},
Clinton Wang\textsuperscript{1}, 
Linxi Fan\textsuperscript{2}, 
\\
\textbf{Tao Chen\textsuperscript{1}, 
De-An Huang\textsuperscript{2}, 
Ekin Aky\"urek\textsuperscript{1}, 
Anima Anandkumar\textsuperscript{2,3,\textdagger}}, 
\\
\textbf{Jacob Andreas\textsuperscript{1,\textdagger}, 
Igor Mordatch\textsuperscript{4,\textdagger}, 
Antonio Torralba\textsuperscript{1,\textdagger}, 
Yuke Zhu\textsuperscript{2,5,\textdagger}} 
\\ \and
\textsuperscript{1}MIT, \textsuperscript{2}Nvidia, \textsuperscript{3}Caltech, \textsuperscript{4}Google Brain, \textsuperscript{5}UT Austin
\\
Junior authors are ordered based on contributions and senior authors\textsuperscript{\textdagger} are ordered alphabetically.
}

\begin{document}

\maketitle

\input{text/abstract}

\input{text/introduction}

\input{text/related_work}

\input{text/problem_setup}

\input{text/policy_across_envs}

\input{text/experiment_main}

\input{text/experiment_analysis}

\input{text/qualitative}

\input{text/conclusion}

\newpage
\setcitestyle{numbers,square}
\bibliographystyle{abbrv}
\bibliography{references}

\clearpage
\input{text/checklist}

\input{text/appendix}



\end{document}

%% file: packages.tex
\usepackage{booktabs}
\usepackage{colortbl}
\usepackage{comment}

\usepackage{enumitem}
\usepackage{hyperref}
\usepackage[ruled,vlined]{algorithm2e}
\usepackage{algorithmic}

\usepackage{url}            
\usepackage{amsfonts}       
\usepackage{nicefrac}       
\usepackage{microtype}      
\usepackage{xcolor}         
\usepackage{bbm}
\usepackage{amsmath}

\usepackage{dsfont} 

\usepackage{float,wrapfig}

\usepackage{array}
\usepackage{adjustbox}
\usepackage{multirow}
\usepackage{colortbl}
\usepackage{caption}
\usepackage{amsmath}
\usepackage{multicol}
\usepackage{mathtools}

\usepackage{soul}

\usepackage{xspace}



%% file: macros.tex
\makeatletter
\DeclareRobustCommand\onedot{\futurelet\@let@token\@onedot}
\def\@onedot{\ifx\@let@token.\else.\null\fi\xspace}

\def\eg{\emph{e.g}\onedot} 
\def\ie{\emph{i.e}\onedot}

\makeatother

\newcommand{\sect}[1]{Section~\ref{#1}}

\newcommand{\fig}[1]{Figure~\ref{#1}}
\newcommand{\tbl}[1]{Table~\ref{#1}}

\newcommand{\ignore}[1]{}

\definecolor{rowblue}{RGB}{220,230,240}
\definecolor{myorchid}{RGB}{150,10,30}
\definecolor{myblue}{RGB}{10,30,250}
\definecolor{mygreen}{RGB}{10,120,10}

\newcommand{\modelname}[0]{\mbox{\textsc{LID}}\xspace}

%% file: math_commands.tex
\usepackage{amsmath,amsfonts,bm}

\def\eqref#1{equation~\ref{#1}}

\def\1{\bm{1}}

\def\va{{\bm{a}}}

\def\vy{{\bm{y}}}

\DeclareMathAlphabet{\mathsfit}{\encodingdefault}{\sfdefault}{m}{sl}
\SetMathAlphabet{\mathsfit}{bold}{\encodingdefault}{\sfdefault}{bx}{n}

\DeclareMathOperator*{\argmin}{arg\,min}

%% file: text/abstract.tex
\begin{abstract}

Language model (LM) pre-training is useful in many language processing tasks. But can pre-trained LMs be further leveraged for more general machine learning problems? We propose an approach for using LMs to scaffold learning and generalization in general sequential decision-making problems. In this approach, goals and observations are represented as a sequence of embeddings, and a policy network initialized with a pre-trained LM predicts the next action. We demonstrate that this framework enables effective combinatorial generalization across different environments and supervisory modalities. We begin by assuming access to a set of expert demonstrations, and show that initializing policies with LMs and fine-tuning them via behavior cloning improves task completion rates by 43.6\% in the VirtualHome environment. Next, we integrate an active data gathering procedure in which agents iteratively interact with the environment, relabel past ``failed'' experiences with new goals, and update their policies in a self-supervised loop. Active data gathering further improves combinatorial generalization, outperforming the best baseline by 25.1\%. Finally, we explain these results by investigating three possible factors underlying the effectiveness of the LM-based policy. We find that sequential input representations (vs.\ fixed-dimensional feature vectors) and LM-based weight initialization are both important for generalization. Surprisingly, however, the format of the policy inputs encoding (e.g.\ as a natural language string vs.\ an arbitrary sequential encoding) has little influence. Together, these results suggest that language modeling induces representations that are useful for modeling not just language, but also goals and plans; these representations can aid learning and generalization even outside of language processing. \let\footnotetext\relax\footnote{
Project page: \href{https://shuangli-project.github.io/Pre-Trained-Language-Models-for-Interactive-Decision-Making/}{\small{https://shuangli-project.github.io/Pre-Trained-Language-Models-for-Interactive-Decision-Making}}.
Part of this work was done during Shuang's internship at NVIDIA.}

\end{abstract}




%% file: text/introduction.tex
\section{Introduction}

\textbf{Language models} (LMs) play a key role in machine learning approaches to natural language processing tasks \citep{devlin2018bert}.
This includes tasks that are not purely linguistic, and require nontrivial planning and reasoning capabilities~\citep{majumdar2020improving,hill2020human}: for example, instruction following, vision-language navigation, and visual question answering. 
Indeed, some of these tasks are so distant from language modeling that one can ask whether pre-trained LMs can be used as a general framework even for tasks that involve no language at all. If so, how might these capabilities be accessed in a model trained only to process and generate natural language strings?


In this paper, we study these questions through the lens of \textbf{embodied decision-making}, investigating the effectiveness of LM pre-training as a general framework for learning policies across a variety of environments.
We propose \textbf{\modelname}, a framework that uses Pre-Trained \textbf{L}anguage Models for \textbf{I}nteractive \textbf{D}ecision-Making. As shown in \fig{fig:framework} (right), we encode the inputs to a policy---including observations, goals, and history---as a sequence of embeddings.
These embeddings are passed to a policy network initialized with the parameters of a pre-trained LM, which is fine-tuned to predict actions.
This framework is broadly applicable, accommodating goals and environment states represented as natural language strings, image patches, or scene graphs.

\input{fig/framework}

We find that imitation learning using pre-trained LMs as policy initializers improves in-domain performance and enables strong generalization over novel tasks.
For i.i.d.\ training and evaluation tasks, this approach yields 20\% more successful policies than other baseline methods in VirtualHome \citep{puig2018virtualhome}.
For combinatorial generalization to out-of-distribution tasks, \ie tasks involving new combinations of goals, states or objects, LM pre-training confers even more benefits: it improves task completion rates by 43.6\% for novel tasks (see \cref{exp:result_baselines}).
These results hold for a variety of environment representations: encoding states as natural language strings, when possible, improves the data-efficiency of training, but even LMs fine-tuned on random environment encodings generalize combinatorially to new goals and states when trained on large enough datasets.

We further examine how our method may be used in environments where expert data is not available, and agents must instead actively gather data. 
To do this, we integrate an \textbf{A}ctive \textbf{D}ata \textbf{G}athering (\textbf{ADG}) procedure into pre-trained LMs as shown in \cref{fig:ilalgo}.
Our proposed approach to ADG consists of three parts.
First, exploration collects trajectories using a mix of random actions and actions generated by the current policy. 
Exploration is insufficient in this high dimensional problem and most of the trajectories will likely fail to achieve the end goal.
A key insight is that even the failed trajectories contain useful sub-trajectories that solve certain sub-goals, and we relabel these goals in a hindsight relabeling stage.
The relabeled goal describes what was achieved in the extracted sub-trajectory.
The policy update stage samples relabeled trajectories to update the policy.
The active data gathering procedure allows us to train the LM-policy without pre-collected expert data. 
It also outperforms reinforcement learning (RL) methods on embodied decision-making tasks and enables more effective generalization to novel tasks.



Finally, we investigate \emph{why} LID contributes to generalization.  
We hypothesize three possible causes for the effectiveness of LM-based policy initialization:
(1) the use of \emph{language-based input encodings}, and more generally LMs' ability to reason about natural language strings;
(2) the \emph{sequential structure} of transformer inputs, in contrast to the fixed-sized observations used by most policy architectures,
and 
(3) \emph{task-general inductive bias} conferred by weight initialization with LM pretraining.
We investigate (1) by encoding the policy inputs as different types of sequences. Different input encoding schemes have only a negligible impact on the performance: the effectiveness of language modeling is not limited to utilizing natural strings, but in fact extends to arbitrary sequential encodings.
We study (2) by encoding observations with a single vector embedding, thereby removing its sequential structure. This operation significantly degrades the model's performance on novel tasks.
Finally, we investigate (3) by learning the parameters of the policy from scratch. The success rate after removing the pre-trained LM weights drops by $11.2\%$, indicating that LM pretraining provides useful inductive bias for sequence processing even when sequences are not natural language strings.

To summarize, our work has four main contributions:
\vspace{-5pt}
\begin{itemize}[leftmargin=1.2em]
    \item First, we propose to use \textbf{pre-trained LMs as a general scaffold} for interactive decision-making across a variety of environments by converting all policy inputs into sequential data.
    \item Second, we demonstrate that \textbf{language modeling improves combinatorial generalization in policy learning}: initializing a policy with a pre-trained LM substantially improves out-of-distribution performance on novel tasks.
    \item Third, we integrate an \textbf{active data gathering} procedure into the proposed approach to further enable policy learning on environments without using pre-collected expert data.
    \item Finally, we perform several analyses to explain the generalization capabilities of pre-trained LMs, finding that natural strings are not needed to benefit from LM pre-training, but the sequential input encoding and weight pre-training are important.
\end{itemize}
%
These results point to the effectiveness of the proposed framework with pre-trained LMs as a general-purpose framework to promote structured generalization in interactive decision-making.


%% file: fig/framework.tex
\begin{figure*}
\begin{center}
\includegraphics[width=1\linewidth]{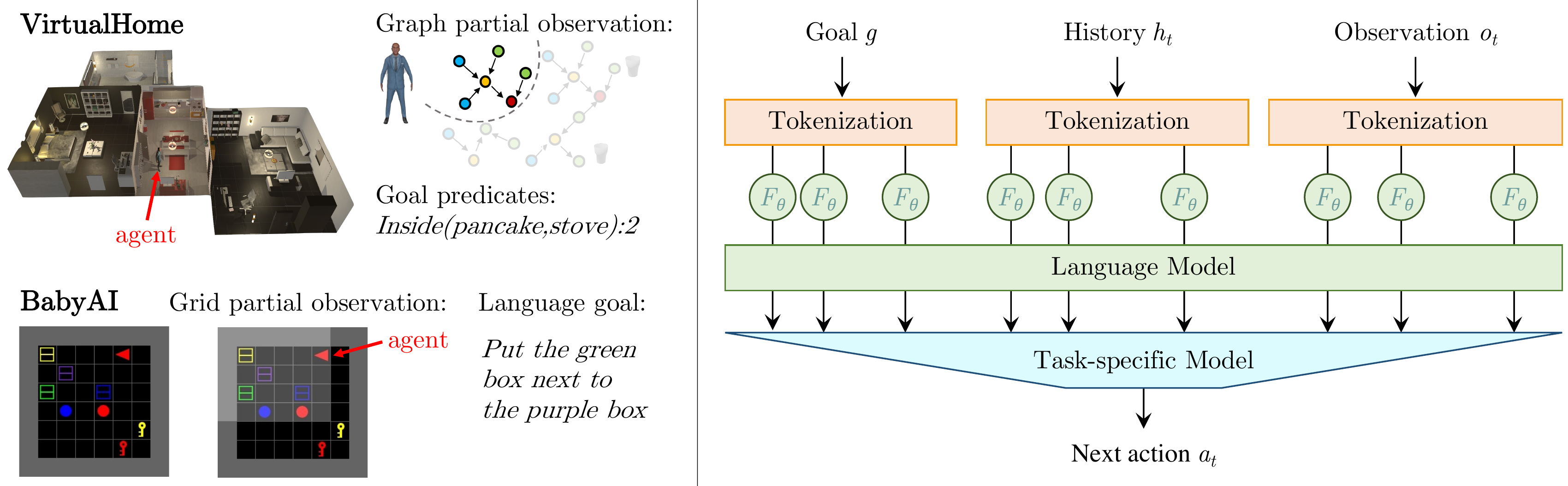}
\end{center}
\vspace{-5pt}
\caption{\small
\textbf{Environments (left):}
Different environments have different types of observations and goals.
\textbf{Our approach (right):} We use pre-trained LMs as a general framework for interactive decision-making by converting policy inputs into sequential data. Such a method enables effective combinatorial generalization to novel tasks.
}
\label{fig:framework}
\vspace{-15pt}
\end{figure*}

%% file: text/related_work.tex
\section{Related Work}
\label{sec:related}
In recent years, word and sentence representations from pre-trained LMs
\citep{peters2018deep,devlin2018bert,radford2018improving} have become ubiquitous in natural language processing~\citep{zhu2020incorporating,platanios2021value}.
Some of the most successful applications of pre-training lie at the boundary of
natural language processing and other domains, as in instruction following \citep{hill2020human} 
and language-guided image retrieval \citep{lu2019vilbert}.

\noindent\textbf{Learning representations of language.}
From nearly the earliest days of the
field, natural language processing researchers observed that
representations of words derived from distributional statistics in large text
corpora serve as useful features for downstream tasks \citep{deerwester1990indexing, dumais2004latent}. The earliest versions of these representation learning schemes focused on isolated word forms
\citep{mikolov2013distributed, pennington2014glove}. However, recent years have seen a number of techniques for training (masked or autoregressive) language models to produce contextualized word representations (which incorporate information neighboring words in sentences
and paragraphs) via a variety of masked-word prediction objectives \citep{devlin2018bert, yang2019neurips}.

\noindent\textbf{Applications of pre-trained LMs.} 
LMs can be fine-tuned to perform language processing tasks other than language modeling by casting those tasks as word-prediction problems. 
Successful uses of representations from pre-trained models include syntactic parsing \citep{kitaev2018multilingual} and language-to-code translation
\citep{wang2019rat}; successful adaptations of LM prediction heads include machine translation \citep{zhu2020incorporating}, sentiment classification \citep{brown2020language} and style transfer \citep{keskar2019ctrl}.
A number of tasks integrate language and other modalities, including visual question answering and image captioning \citep{yang2020bert}. 
Recent works find that image representations can be injected directly into LMs' embedding layers \citep{tsimpoukelli2021multimodal}.

\noindent\textbf{Policy learning and LM.}
Traditional policy learning methods, such as PPO~\citep{schulman2017proximal}, DQN~\citep{mnih2013playing}, DDPG~\citep{lillicrap2015continuous}, A3C~\citep{mnih2016asynchronous}, perform well on playing tasks on Atari, OpenAI gym~\citep{1606.01540}, and MuJoCo~\citep{todorov2012mujoco}. Some of them might fail to solve more challenging tasks on embodied environments~\citep{puig2018virtualhome,shen2020igibson}.
Several recent papers~\citep{reid2022can,jang2022bc,huang2022language} propose to use LM for policy learning.
Frozen Pretrained Transformer (FPT)~\citep{lu2021pretrained}
demonstrates that pre-trained LMs require very little fine-tuning to \emph{match}
the performance of task-specific models on several image classification and
numerical sequence processing tasks. 
Semi-Supervised Skill Learning with Latent Language (SL)$^3$~\citep{sharma22sl3} shows that LMs can serve as an effective backbone for hierarchical policies that express plans as natural language strings \cite{andreas19l3, jacob21llp}.
In this paper, we focus on building a general framework for decision-making tasks using pre-trained LMs, even when language is not provided as an input or output.

%% file: text/problem_setup.tex
\section{Decision-Making and Language Modeling}
\subsection{POMDPs and Policy Learning}
We explore the application of LMs to general sequential decision-making tasks in partially observed environments.
These tasks may be formalized as partially observable Markov decision processes (POMDPs).
A POMDP is defined by a set of states, a set of observations, a set of actions, and a transition model $\mathcal{T}(s_{t+1} | s_t, a_t)$ that maps the current state and action to the next state.
Importantly, in a POMDP setting, the observation $o_t$ only captures a portion of the underlying state $s_t$, and an optimal decision-making strategy (a \textbf{policy}) must incorporate both the current observation and the history of previous observations and actions. In our experiments, policies are parametric models $\pi_\phi(a_t | g, h_t, o_t)$ that output the probability of an action given the goals $g$, history information $h_t = \{o_1, a_1, \cdots, o_{t-1}, a_{t-1}\}$, and partial observations $o_t$ of the current state $s_t$. 

In \cref{fig:framework} (right), we show a high-level overview of the proposed method. We first convert all policy inputs into a sequence and provide them as input to a transformer encoder. Representations from this encoder model are then passed to a task-specific decoder that predicts actions.
%
We collect a dataset of $N$ training trajectories $\mathcal{D} = \{d^{i}\}_{i=1}^N$, where each trajectory consists of a goal and a sequence of observations and actions: $d^{i} = \{g^{i}, o_1^{i}, a_1^{i}, \cdots, o_{T_i}^{i}, a_{T_i}^{i}\}$, where $T_i$ is the length of the trajectory. 
We then train the policy to maximize the probability of actions we want to achieve $\va^{i}=\{a_1^{i},\ldots,a_{T_i}^{i}\}$ across trajectories using the cross-entropy loss:
\begin{equation}
\phi^* = \argmin_{\phi} \left( - \sum_{i=1}^N \sum_{t=1}^{T_i} \ln \pi_\phi (a_{t}^{i} | g^{i}, h_t^{i}, o_t^{i}) \right).
\label{eq:policy} 
\end{equation}

\subsection{Language models as policy initializers}
\label{sect:lm_policy}
\vspace{-3pt}
Our experiments focus on \textbf{autoregressive}, \textbf{transformer-based LMs} \citep{vaswani2017attention}.
These models are trained to fit a distribution over a text sequence $\vy = \{y_i\}_{i=1}^n$ via the chain rule
$p(\vy) = p(y_1) \prod_{i=2}^n p(y_i \mid y_1, \ldots, y_{i-1})$.
Each term on the right hand side is parameterized by a transformer network, which accepts the conditioned tokens as input. Each token passes through a learned embedding layer $F_\theta$, then the full conditioned sequence is fed into the LM.
In our work, we use a standard LM, GPT-2, to process the input sequence rather than to predict future tokens.

Both POMDP decision-making and language modeling are naturally framed as sequence prediction tasks, where successive words or actions/observations are predicted based on a sequence of previous words or actions/observations. This suggests that pre-trained LMs can be used to initialize POMDP policies by fine-tuning them to model high-reward or expert trajectories, as described below.

%% file: text/policy_across_envs.tex
\section{Approach}
We evaluate the effectiveness of pre-trained LMs in solving decision-making tasks across environments.
We use \textbf{BabyAI} \citep{hui2020babyai} and \textbf{VirtualHome} \citep{puig2018virtualhome} to evaluate the proposed method.
While both environments feature complex goals, the nature of these goals, as well as the state and action sequences that accomplish them, differ substantially across environments (\cref{fig:framework} (left)).


\subsection{Policy Network}
\label{sec:policy_network}
We first examine whether pre-trained LMs provide effective initializers when states and action histories are represented as natural language strings.
We encode the inputs to the policy---including observations, goals, and action histories---as sequences of words.
These word sequences are passed to the LM (using its pre-trained word embedding layer $F_{\theta}$) and used to obtain contextualized token representations. Token representations are averaged and used to predict actions.
We design a policy network following the general policy framework proposed in \cref{fig:framework}.

\textbf{Environment encodings in VirtualHome.}
In VirtualHome, each goal consists of a sequence of predicates and multiplicities, and is translated into a templated English sentence (\eg ``\texttt{Inside(apple, fridge):2}'' becomes ``put two apples inside the fridge'').
%
To encode the agent's partial observation, we extract a list of currently visible objects, their states (\eg ``open, clean''), and 3D world coordinates. We use a fully-connected layer to encode the 3D information and generate a feature representation of each object in the observation. 
To encode history, we store information about all previous actions and convert them into templated English sentences (\eg ``I have put the plate on the kitchen table and the apple inside the fridge'').

\textbf{Environment encodings in BabyAI.}
The observation by default is a $7 \times 7$ grid.
We convert the observation into $7 \times 7$ text descriptions, \eg ``purple ball'', ``grey wall'', ``open door'', and combine them into a long sentence.
We then convert the history actions into text descriptions, \eg ``turn left'' and ``go forward''.
We combine the language instruction (without modification) with the observation and history text descriptions, and feed them to the pre-trained LM.

We note that the policy network described above does not strictly require that these encodings take the form of natural language strings---other encodings of the environment as a sequence also work (see \cref{sect:pre-trained-learned-encoding}).
This framework could be also generalized to support pixel-based observations using discretization schemes like the one employed in the Vision Transformer \citep{dosovitskiy2020image}.

\textbf{Action prediction.}
We pool LM outputs into a ``context representation'' that is used to predict the next action. 
In training, we maximize the probabilities of demonstrated actions.
In inference, we select the valid action with the highest probability.
See \textbf{Appendix \ref{apx:model_architecture}} for details.

VirtualHome and BabyAI have quite different observation spaces, action spaces, and goal spaces; however, we show that embedding policy inputs as sequences and utilizing the pre-trained LM as a policy initializer, enables effective generalization to novel tasks on both environments.
We note that \modelname is not limited to VirtualHome and BabyAI, but is straightforwardly applicable to other embodied environments, such as ALFRED~\citep{shridhar2020alfred} and iGibson~\citep{shen2020igibson}.

\subsection{Training}
\vspace{-3pt}
We first examine \modelname through imitation learning on data collected by experts in \cref{sec:imitation-learning}. We then show that integrating an active data gathering procedure into \modelname enables policy learning without using expert data in \cref{sec:active-data-collection}.
We use VirtualHome as an example to explain the data gathering.

\subsubsection{Policy Learning with Expert Data}
\label{sec:imitation-learning}
The policy model is first initialized from a pre-trained LM and then fine-tuned on data collected by experts.
We build on the VirtualHome environment to collect a set of expert trajectories using regression planning \citep{korf1987planning} and create a \textbf{VirtualHome-Imitation Learning dataset}.
Given a task described by goal predicates, the planner generates an action sequence to accomplish this task (See \textbf{\cref{apx:expert_data_collection}}).
The planner has access to privileged information, such as information about the pre-conditions and effects of each action, allowing an agent to robustly perform tasks in partially observable environments and generate expert trajectories for training and evaluation.

\subsubsection{Policy Learning with Active Data Gathering}
\label{sec:active-data-collection}
\input{fig/2hindsight}

Collecting expert data is sometimes challenging. It may require privileged information of the environment or human annotations, which can be time-consuming and difficult to scale.
A promising way to scale up supervision is Hindsight Experience Replay (HER)~\cite{andrychowicz2017hindsight}, which allows agents to learn from orders of magnitude more data without supervision.
However, existing HER methods~\citep{ghosh2019learning} focus on simple tasks with small state/action space and full observability. 
They cannot tackle more complicated embodied decision-making tasks, requiring nontrivial planning and reasoning or natural language understanding. 
\modelname with the active data gathering (\textbf{\modelname-ADG}) can be used in solving tasks in such environments.

As shown in \cref{fig:ilalgo}, \modelname-ADG consists of three stages, \ie \textbf{exploration}, \textbf{hindsight relabeling}, and \textbf{policy update}.
The key idea is to gradually improve the task success rate by asking the agent to iteratively explore the environment, relabel failure samples, and update its policy using imitation learning.
In the \textbf{exploration} stage, we first randomly sample a goal and an initial state. 
We then use a mix of random actions and actions generated by the current policy $\pi_{\phi}(a_{t} | g,h_t,o_t)$ to obtain the next action.
We repeat this process until this episode ends.
We collect $M$ trajectories and store them in the replay buffers.
The generated actions in the early stages rarely complete the given task. 
However, even the failed trajectories contain useful sub-trajectories that solve certain sub-goals.
In the \textbf{hindsight relabeling} stage, we extract useful sub-trajectories and relabel a goal $g'$ for each of them.
We design a goal relabel function $f_l$ that generates a goal based on the sequence of observations and actions using hand-designed templates. 
In practice, we implement the goal relabel function as a program (see \textbf{\cref{apx:active_data_gathering}}).
The \emph{hindsight relabeling} stage allows sample-efficient learning by reusing the failure cases.
During \textbf{policy update}, the agent samples the data from the replay buffers and updates its policy network $\pi_{\phi}$.

By interleaving the exploration, hindsight relabeling, and policy update, \modelname-ADG can gradually improve the policy without requiring pre-collected expert data. 
In embodied environments with large action spaces, sparse rewards, and long-horizon planning, RL methods often struggle to obtain stable policy gradients during training. 
Our method enables sample-efficient learning from the sparse rewards by relabeling new goals for the bad samples that the agent fails to achieve.
In addition, \modelname-ADG leverages the stability of supervised learning in the \emph{policy update} stage, enabling it to outperform RL approaches on a wide range of decision-making tasks.


%% file: fig/2hindsight.tex
\begin{wrapfigure}{r}{0.6\textwidth}
\vspace{-14pt}
\begin{center}
\includegraphics[width=1\linewidth]{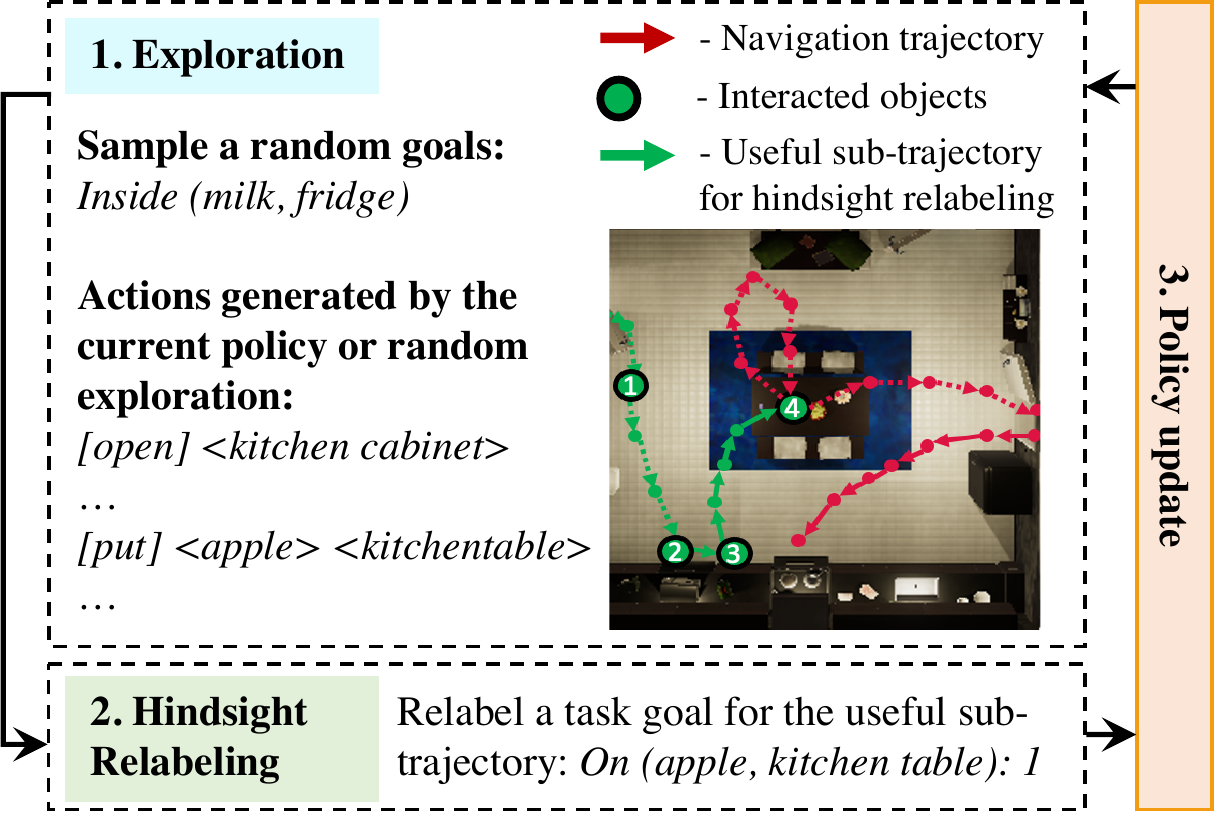}
\end{center}
\vspace{-5pt}
\caption{\small \textbf{\modelname with the active data gathering procedure.}
By iteratively repeating the exploration, hindsight relabeling, and policy update, \modelname with active data gathering can learn an effective policy without using pre-collected expert data.}
\label{fig:ilalgo}
\vspace{-5pt}
\end{wrapfigure}

%% file: text/experiment_main.tex
\section{Experiment Setup}
\vspace{-5pt}
We evaluate the proposed method and baselines on VirtualHome and BabyAI.

\vspace{-3pt}
\subsection{VirtualHome}
\label{sec:exp_setup_vh}
VirtualHome is a 3D embodied environment featuring partial observability, large action spaces, and long time horizons. 
We evaluate policies' performance from three aspects:
(1) performance on in-distribution tasks;
(2) generalization to novel scenes;
and (3) generalization to novel tasks.

\textbf{In-Distribution.}
The predicate types and their counts in the goal are randomly sampled from the same distribution as the training data. 
The objects are initially placed in the environment according to common-sense layouts (\eg plates appear inside the kitchen cabinets rather than the bathtub).

\textbf{Novel Scenes.}
The objects are placed in random positions in the initial environment without common-sense constraints (\eg apples may appear inside the dishwasher).

\textbf{Novel Tasks.}
The components of all goal predicates are never seen together during training (\eg both plates and fridges appear in training goals, but \texttt{Inside(plate, fridge)} only appears in the test set.
(See \textbf{\cref{apx:test_subsets}} for more details.)

We evaluate the success rates of different methods on each test set.
A given episode is scored as successful if the policy completes its entire goal within the maximum allowed steps of the environment.
On each of the 3 test subsets, we use 5 different random seeds and test $100$ tasks under each seed.
Thus there are $1500$ examples used to evaluate each model.

\vspace{-3pt}
\subsection{BabyAI}
\vspace{-3pt}
BabyAI is a 2D grid world environment for instruction following. 
Observations in BabyAI are $7 \times 7 \times 3$ grids describing a partial and local egocentric view of the state of the environment. 
We evaluate the methods on four representative tasks: \emph{GoToRedBall}, \emph{GoToLocal}, \emph{PickupLoc}, and \emph{PutNextLocal}.
Performing well on the test set requires the models to generalize to new environment layouts and goals, resulting in new combinations of tasks not seen in training. 
For each method, we compute success rates over $500$ episodes on each task.




\section{Experiments}
\label{sec:experiments_main}
\vspace{-3pt}
We first show results of the proposed method and baselines for embodied decision-making tasks using expert data in \sect{sec:exp_offline_data}. We then show our results when using actively gathered data in \sect{sec:exp_active_data}.

\vspace{-3pt}
\subsection{Embodied Decision Making with Pre-trained Language Model (\modelname)}
\label{sec:exp_offline_data}

\subsubsection{Results on VirtualHome}

We evaluate the following methods:

\textbf{\modelname-Text (Ours)} is the proposed method that converts all environments inputs into text descriptions. The pre-trained LM is fine-tuned for decision-making (conditioned on goals, observations, and histories) as described in \cref{sec:policy_network}.

\textbf{Recurrent Network}. We compare our method with a recurrent baseline using an LSTM~\citep{hochreiter1997long} to encode the history information. The hidden representation from the last timestep, together with the goal and current observation, are used to predict the next action.

\textbf{MLP} and \textbf{MLP-1}.
We perform additional comparisons with baselines that do not use recurrent networks or pre-trained LMs. 
\emph{MLP} and \emph{MLP-1} take the goal, histories, and the current observation as input and send them to the multilayer perceptron neural network (MLP) to predict actions.
\emph{MLP-1} has three more average-pooling layers than \emph{MLP} that average the features of tokens in the goal, history actions, and the current observation, respectively, before sending them to the MLP layer.

\textbf{Quantitative results}.
Each method is trained on $20K$ demos from the VirtualHome-Imitation Learning dataset, and then evaluated on the three test subsets: \textbf{In-Distribution}, \textbf{Novel Scenes}, and \textbf{Novel Tasks}. In \fig{exp:result_baselines}, \emph{\modelname-Text (Ours)}, which initializes the policy with a pre-trained LM, has higher success rates than other methods.
This difference is most pronounced in the \textbf{Novel Tasks} setting, where test tasks require combinatorial generalization across goals that are never seen during training. Here, \emph{\modelname-Text (Ours)} dramatically ($43.6\%$) improves upon all baselines. 
Such combinatorial generalization is necessary to construct general purpose agents, but is often difficult for existing approaches. Our results suggest that pre-trained LMs can serve as a computational backbone for combinatorial generalization. 

\subsubsection{Results on BabyAI}
We use the standard training and test data provided by \citep{hui2020babyai}. 
In BabyAI, performing well on unseen test tasks with new environment layouts and goals requires combinatorial reasoning.
In \cref{exp_tbl:babyai_baselines}, we report the success rate of models trained on different number of demos.
\textbf{BabyAI-Ori}~\citep{hui2020babyai} is the method used in the original paper. \textbf{\modelname-Text (Ours)} is the proposed method that converts policy inputs into a text sequence.
Given enough training data, \ie 10K demos, both methods achieve high success rates, but \emph{\modelname-Text (Ours)} outperforms BabyAI-Ori with less training data, indicating the proposed method improves sample efficiency when generalizing to novel tasks.

\input{fig/11baselines}

\subsection{Pre-trained Language Model with Active Data Gathering (\modelname-ADG)}
\label{sec:exp_active_data}
We compare \textbf{\modelname-ADG}, the proposed LM framework for decision-making using actively gathered data (\cref{sec:active-data-collection}), to a variety of baselines that do not use pre-collected expert data on VirtualHome.

\textbf{Random.} The agent selects the next action randomly from the valid action space at that state.
\textbf{Goal-Object.}
The agent randomly selects an object that in the goal and in the valid action space to interact with. For example, given a goal of ``\texttt{Inside(apple, fridge):1}'', this baseline might choose ``grab apple'', ``open fridge'', or other actions containing ``apple'' or ``fridge''.
\textbf{Online RL.} 
We compare with PPO~\citep{schulman2017proximal}, one of the most commonly used online RL methods. For fair comparison, we equip PPO with the same main policy network as the proposed method. Our implementation is based on Stable Baselines3~\citep{stable-baselines3}.
\textbf{Hindsight Experience Replay.} 
We compare with DQN+HER used in \citep{andrychowicz2017hindsight} and modify its main policy network to be the same as the proposed method.

\textbf{Quantitative results}.
We compare \modelname-ADG with baselines on VirtualHome in \cref{exp:baselines}.
Each experiment is performed 5 times with different random seeds. 
The \textbf{Random} baseline is always 0, indicating the tasks in VirtualHome cannot be easily solved by a random policy.
\textbf{Goal-Object} is better than \emph{Random} because \emph{Goal-Object} has access to objects in the goal and it samples actions from a much smaller action space.
The online RL baseline, \textbf{PPO}, fails to solve tasks in VirtualHome featured by partially observation, large state/action space, and long-term horizon.
\textbf{DQN+HER} works well on simple tasks on 2D environments, but they cannot tackle VirtualHome tasks neither, requiring nontrivial planning and reasoning.
%
\modelname-ADG does not require expert data and can solve the complicated tasks in 3D embodied environments which cannot be easily achieved using RL.
\footnote{Note that the results of \modelname-Text in \cref{exp:result_baselines} and results of \modelname-ADG in \cref{exp:baselines} are not directly comparable because the difficulty level of the evaluated tasks are different. See \textbf{\cref{apx:test_subsets}} for more details.}

\textbf{Policy initializer and data provider.}
\modelname-ADG can further be used to initialize the weights for fine-tuning RL policies and to gather data for offline learning.
As shown in \cref{exp:baselines}, directly training RL, \eg PPO, fails to solve tasks in VirtualHome.
However, after using the policy trained by \modelname-ADG to initialize the PPO policy, we may effectively learn an interactive policy with good performance.
In \cref{exp:provider}, \textbf{PPO (\modelname-ADG Init)} is initialized from \modelname-ADG and further fine-tuned to solve the tasks in VirtualHome.
After initialization, PPO improves its success rate by $53.7 \%$ on the \emph{In-Distribution} setting (See PPO results in \cref{exp:baselines} and \cref{exp:provider}).
In addition, \modelname-ADG can provide data for offline learning.
\modelname-ADG saves the relabeled data in replay buffers.
We train Decision Transformer (DT)~\citep{chen2021decision} using the data collected by \modelname-ADG. See \textbf{DT (\modelname-ADG Data)} in \cref{exp:provider}.




\input{fig/12sindsight}


%% file: fig/11baselines.tex
\begin{figure*}[t]
\centering
\hfill
\begin{minipage}{0.5\textwidth}
    \centering
    \small
    \includegraphics[width=1\textwidth]{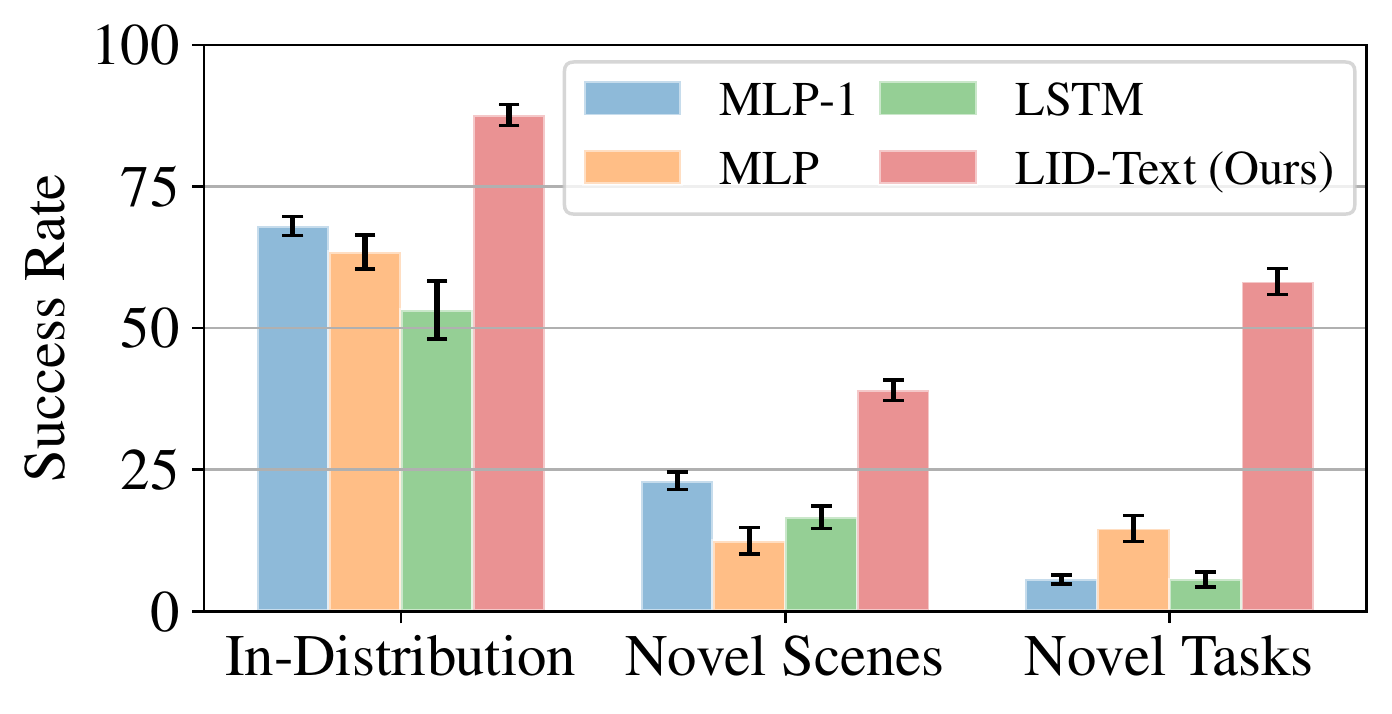}
    \vspace{-12pt}
    \caption{\small{\textbf{Comparisons of the proposed method and baselines on VirtualHome.} All the methods are trained on expert data using imitation learning. 
    \emph{MLP-1}, \emph{MLP}, and \emph{LSTM} are baselines without using the pre-trained LM. 
    The proposed method, \emph{\modelname-Text (Ours)}, outperforms all baselines.
    }}
\label{exp:result_baselines}
\end{minipage}
\hfill \hfill
\begin{minipage}{0.48\textwidth}
    \centering
    \small
      \centering
      \addtolength{\tabcolsep}{-3.9pt}
      \scalebox{0.8}{
      \begin{tabular}{llcccccc}
    \toprule
    \bf Tasks & \bf Methods & \multicolumn{5}{c}{\bf Number of Demos} \\
     \cmidrule(lr){3-7}
    & &  \bf 100 & \bf 500 & \bf 1K & \bf 5K & \bf 10K \\
    
    \midrule
    
    \multirow{2}{*}{\bf GoToRedBall} 
    & \bf BabyAI-Ori \citep{hui2020babyai} & 81.0 & 96.0 & 99.0 & 99.5 & 99.9 \\
    & \bf \modelname-Text (Ours) & \bf 93.9 & \bf 99.4 & \bf 99.7 & \bf 100.0 & \bf 100.0 \\
    
    \midrule
    
    \multirow{2}{*}{\bf GoToLocal} 
    & \bf BabyAI-Ori \citep{hui2020babyai} & 55.9 & 84.3 & 98.6 & \bf 99.9 & 99.8 \\
    & \bf \modelname-Text (Ours) & \bf 64.6 & \bf 97.9 & \bf 99.0 & 99.5 & 99.5 \\
    
    \midrule
    
    \multirow{2}{*}{\bf PickupLoc} 
    & \bf BabyAI-Ori \citep{hui2020babyai} & 28.0 & 58.0 & 93.3 & 97.9 & \bf 99.8 \\
    & \bf \modelname-Text (Ours) & \bf 28.7 & \bf 73.4 & \bf 99.0 & \bf 99.6 & \bf 99.8 \\
    
    \midrule
    
    \multirow{2}{*}{\bf PutNextLocal} 
    & \bf BabyAI-Ori \citep{hui2020babyai} & \bf 14.3 & 16.8 & 43.4 & 81.2 & 97.7 \\
    & \bf \modelname-Text (Ours) & 11.1 & \bf 93.0 & \bf 93.2 & \bf 98.9 & \bf 99.9 \\
    
    \bottomrule
    \end{tabular}
    }
    \vspace{-5pt}
    \captionof{table}{\small 
         \textbf{Success rates on BabyAI tasks.} All the methods are trained on offline expert data using imitation learning. 
         \emph{\modelname-Text (Ours)} outperforms BabyAI-Ori, the method used in the original paper \citep{hui2020babyai}.
         }
    \label{exp_tbl:babyai_baselines}
    \end{minipage}%
\vspace{-10pt}
\end{figure*}

%% file: fig/12sindsight.tex
\begin{figure*}[t]
\centering
\hfill
\begin{minipage}{0.47\textwidth}
    \centering
    \small
    \setlength{\tabcolsep}{0.5em}
      \centering
      \addtolength{\tabcolsep}{-3.5pt}
      \scalebox{0.8}{
      \begin{tabular}{lccccc}
    \toprule
    & \bf {In-Distribution} & \bf {Novel Scenes} & \bf \shortstack{Novel Tasks} \\
    \midrule
    
    \bf Random  & 0.0 $\pm$ 0.0 & 0.0 $\pm$ 0.0 & 0.0 $\pm$ 0.0  \\
    \bf Goal-Object & 0.8 $\pm$ 0.5 & 0.0 $\pm$ 0.0 & 0.4 $\pm$ 0.4 \\
    \bf PPO & 0.0 $\pm$ 0.0 & 0.0 $\pm$ 0.0 & 0.0 $\pm$ 0.0 \\
    \bf DQN+HER & 0.0 $\pm$ 0.0 & 0.0 $\pm$ 0.0 & 0.0 $\pm$ 0.0 \\
    \bf \modelname-ADG (Ours) & \bf 46.7 $\pm$ 2.7 & \bf 32.2 $\pm$ 3.3 & \bf  25.5 $\pm$ 4.1  \\
    
    \bottomrule
  \end{tabular}
    }
    \vspace{-5pt}
    \captionof{table}{\small{\textbf{Comparisons of methods without using expert data on VirtualHome.} \emph{\modelname-ADG (Ours)} is the only successful approach.}}
  \label{exp:baselines}
\end{minipage}
\hfill \hfill
\begin{minipage}{0.5\textwidth}
    \small
      \centering
      \setlength{\tabcolsep}{0.5em}
      \addtolength{\tabcolsep}{-3.3pt}
      \scalebox{0.8}{
    \begin{tabular}{lccccc}
    \toprule
    & \bf {In-Distribution} & \bf {Novel Scenes} & \bf \shortstack{Novel Tasks} \\
    \midrule
    
    \bf \modelname-ADG (Ours) & 46.7 $\pm$ 2.7 & \bf 32.2 $\pm$ 3.3 & 25.5 $\pm$ 4.1  \\
    \bf PPO (\modelname-ADG Init) & \bf 53.7 $\pm$ 3.5 & 30.2 $\pm$ 3.4 & \bf 27.8 $\pm$ 2.7 \\
    \bf DT (\modelname-ADG Data) & 42.4 $\pm$ 1.5 & 21.6 $\pm$ 2.48 & 16.8 $\pm$ 1.0 \\
    \bottomrule
    \end{tabular}
    }
    \captionof{table}{\small The proposed method with active data gathering, \modelname-ADG (Ours), can be used as an policy initializer for online RL or a data provider for offline RL.}
    \label{exp:provider}
\end{minipage}
\vspace{-10pt}
\end{figure*}

%% file: text/experiment_analysis.tex
\section{Analysis: Understanding the Sources of Generalization}
\label{sect:pre-trained-learned-encoding}
The pre-trained LM policy, fine-tuned on either expert data or actively gathered data, exhibits effective combinatorial generalization.
Is this simply because LMs are effective models of relations between natural language descriptions of states and actions \cite{ammanabrolu2018playing}, or because they provide a more general framework for combinatorial generalization in decision-making? We hypothesize and investigate three possible factors to understand the sources of such combinatorial generalization.
We use policies trained on the expert data as an example to explain the experiments.

\subsection{Input Encoding Scheme}
\label{sect:input_encoding}
We first hypothesize that converting environment inputs into natural language contributes to the combinatorial generalization as the LMs are trained on language data.
We explore the role of \emph{natural language} by investigating three alternative ways of encoding policy inputs to our model without using natural language strings: two in VirtualHome, and one in BabyAI. BabyAI results are in \cref{apx:convolutional_encoding_babyai}.

\textbf{Index encoding in VirtualHome.}
Rather than natural language strings, \emph{\modelname-Index (Ours)} converts policy inputs into integer indices.
\emph{\modelname-Index (Ours)} retains the discrete, serial format of the goal, history, and observation, but replaces each word with an integer, and replaces the embedding layer from the pre-trained LM with a new embedding layer trained from scratch. 
For example, \emph{grab apple} is mapped to (5,3) based on the positions of \emph{grab} and \emph{apple} in the vocabulary set.

\textbf{Unnatural string encoding in VirtualHome.} \emph{\modelname-Unnatural (Ours)} replaces the \emph{natural language} tokens (e.g.\ converting the goal ``\texttt{On(fork, table):1}'' to \emph{put one fork on the table}) with random ones (e.g.\ converting \texttt{On(fork, table)} to \emph{brought wise character trees fine yet}). This is done by randomly permuting the entire vocabulary, mapping each token to a new token. 
Such a permutation breaks the semantic information in natural strings.
%

\emph{\modelname-Index (Ours)} and \emph{\modelname-Unnatural (Ours)} have the same policy network as \emph{\modelname-Text (Ours)}. All are fine-tuned on the expert data.
The averaged results using 5 different random seeds on the Novel Tasks setting are reported in \cref{exp:learnedemb}.
Given few training data, e.g. 100 demos, all the models perform poorly, with success rates lower than $10\%$. 
\emph{\modelname-Text (Ours)} achieves higher success rates than \emph{\modelname-Index (Ours)} and \emph{\modelname-Unnatural (Ours)} when dataset size increases, \eg \emph{\modelname-Text (Ours)} is around $4\%$ higher than \emph{\modelname-Index (Ours)} and \emph{\modelname-Unnatural (Ours)} with 500 training demos.
When the training dataset is further enlarged, \eg 20K demos, success rates of all approaches reach similar performance.
This result indicates that the effectiveness of pre-trained LMs in compositional generalization is not unique to natural language strings, but can be leveraged from arbitrary encodings, although adapting the model to arbitrary encodings may require more training data.

\input{fig/tokenization-vh}

\subsection{Sequential Input Representation}
\input{fig/3embedding}

Next, we explore whether generalization requires the sequential processing mechanisms in transformer-based LMs.
We investigate whether the LM pre-trained policy will still be effective when the input encoding is not sequential.
\textbf{No-Seq} encodes the goal as a single vector by averaging all goal embeddings. History and observation features are obtained in the same way.
All features are then sent to the pre-trained LM to predict actions.
As shown in \cref{exp:noseq}, removing sequential structure significantly hurts performance on \emph{Novel Tasks}. \emph{No-Seq} achieves good performance on test tasks that are closer to training tasks, but cannot generalize well to more challenging unseen tasks. Thus, combinatorial generalization in pre-trained LMs may be attributed in part to transformers' ability to process sequential input representations effectively.


\subsection{Favorable Weight Initialization}
Finally, we investigate if the favorable weight initialization from LM pre-training enables effective generalization of the proposed model.
%
\textbf{No-Pretrain} does not initialize the policy using the pre-trained LM, but instead training the policy on the expert data from scratch.
In \cref{exp:noseq}, we find that removing the pre-trained weights can fit the in-domain data and thus performs well on the \emph{In-Distribution} setting.
However, its success rate is $11.2\%$ lower than the proposed model on the \emph{Novel Tasks} setting, indicating the pre-trained weights are important for effective generalization, but not necessary for effective data fitting.
We further test a baseline, \textbf{No-FT}, that keeps the pre-trained weights of the language model but freezes them while training the rest model on our expert data. Freezing the pre-trained weights without fine-tuning significantly hurts the performance on both settings, suggesting that fine-tuning of the transformer weights is essential for effective combinatorial generalization.

Together, these results suggest that sequential input representations (vs.\ fixed-dimensional feature vectors) and favorable weight initialization are both important for generalization, however, the input encoding schemes (e.g.\ as a natural language string vs.\ an arbitrary encoding scheme) has little influence.
These results point to the potential broader applicability of pre-trained LMs as a computational backbone for compositional embodied decision making, where arbitrary inputs, such as language, images, or grids, may be converted to sequential encodings. 


%% file: fig/tokenization-vh.tex
\begin{table}
\caption{\small{
  \textbf{Success rates of policies trained with different input encodings in the \emph{Novel Tasks} setting on VirtualHome.} The text encoding is the most sample-efficient, but all models converge to similar performance given sufficient training data.
  }
  }
\label{exp:learnedemb}
\small
  \centering
  \scalebox{0.85}{
  \begin{tabular}{lcccccccc}
    \toprule
    \bf Methods & \multicolumn{6}{c}{\bf Number of Demos} \\
    \cmidrule(lr){2-7}
    & \bf 100 & \bf 500 & \bf 1K & \bf 5K & \bf 10K & \bf 20K \\
    
    \midrule
    
    \bf \modelname-Text (Ours) & \bf 8.8 $\pm$ 1.4 & \bf 22.2 $\pm$ 1.7 & 26.8 $\pm$ 1.0 & 46.0 $\pm$ 1.0 & \bf 58.2 $\pm$ 1.2 & 58.2 $\pm$ 1.6 \\
    \bf \modelname-Index (Ours) & 6.4 $\pm$ 0.6  & 18.0 $\pm$ 3.8 & 18.8 $\pm$ 1.0 & 45.5 $\pm$ 2.1 & 54.6 $\pm$ 0.8 & 57.8 $\pm$ 0.9 \\
    \bf \modelname-Unnatural (Ours) &  6.8 $\pm$ 1.3 & 18.6 $\pm$ 2.1 & \bf 27.0 $\pm$ 1.1 & \bf 47.2 $\pm$ 1.7 & 55.8 $\pm$ 0.8& \bf 58.8 $\pm$ 0.9 \\
    \bottomrule
  \end{tabular}
  }
\vspace{-10pt}
\end{table}

%% file: fig/3embedding.tex

\begin{wraptable}{r}{6.5cm}
\vspace{-10pt}
  \caption{\small \textbf{Experiments on sequential inputs and weight initialization.} 
  Fine-tuning the pre-trained weights and the usage of sequential encoding are important for combinatorial generalization.}
  \vspace{-5pt}
  \label{exp:noseq}
  \small
  \centering
  \setlength{\tabcolsep}{0.3em}
  \scalebox{1}{
  \begin{tabular}{lccccc}
    \toprule
    & \bf {In-Distribution} & \bf \shortstack{Novel Tasks} \\
    \midrule
    \bf \modelname-Text (Ours) & 87.6 $\pm$ 1.9 & \bf 58.2 $\pm$ 2.3 \\
    \bf No-Seq & 74.0 $\pm$ 2.3 & 2.0 $\pm$ 0.6 \\
    \bf No-Pretrain & \bf 90.8 $\pm$ 2.0 & 47.0 $\pm$ 2.8 \\
    \bf No-FT & 51.2 $\pm$ 4.5 & 17.0 $\pm$ 2.9 \\
    \bottomrule
  \end{tabular}
  }
\end{wraptable}

%% file: text/qualitative.tex
\input{fig/trajectory}

\section{Qualitative Results}
In \fig{fig:result}, we show examples of \emph{\modelname-Text (Ours)} completing tasks in VirtualHome and BabyAI. 
We show two successful examples from VirtualHome on the \emph{In-Distribution} and \emph{Novel Tasks} settings, and two successful examples from BabyAI on solving the \emph{GoToLocal} and \emph{PickupLoc} tasks.
We only show short trajectories or extract a sub-trajectory for saving space. 

\textbf{Failure case analysis.}
In \fig{apx_fig:qualitative}, we show some failure cases of the proposed method.
We observed two main types of failure cases: grounding error and policy error.
For failures caused by the grounding error, the agent interacts with a wrong object that is not related to the given goal, \eg the agent puts \emph{cutlets} instead of the \emph{salmon} inside the fridge.
For failures caused by the policy error, the agent cannot find the target objects or does not interact with them.
The proposed method that converts policy inputs into sequential encodings and feeds them to the general LM framework can accomplish decision-making tasks efficiently, however, there are still challenging tasks that the policy fails to accomplish.
Larger LMs, \eg GPT-3 \citep{brown2020language}, may improve the success rate of those challenging tasks.

\input{appendix/fig/apx_qualitative}

%% file: fig/trajectory.tex

\begin{figure*}[t]
\begin{center}
\includegraphics[width=1\textwidth]{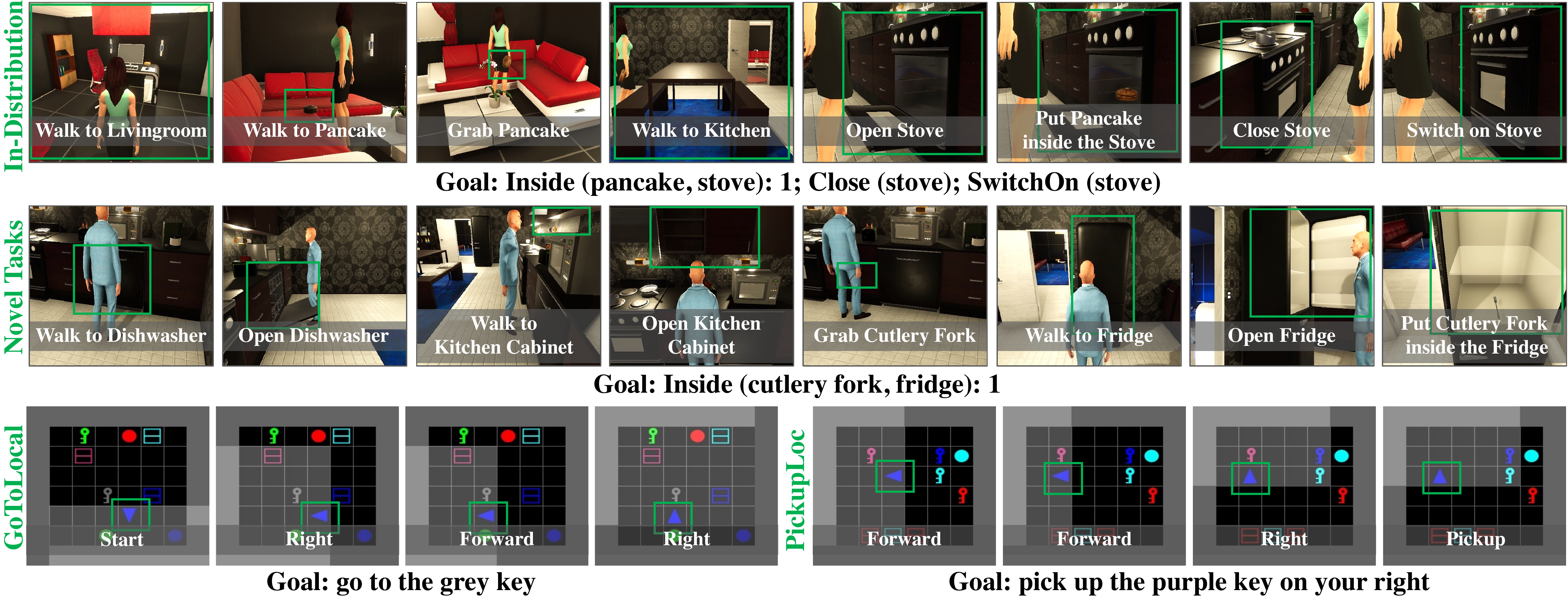}
\end{center}
\vspace{-7pt}
\caption{\small \textbf{Qualitative results of our model on VirtualHome and BabyAI.} 
We only show a sub-trajectory in each example to save space. 
The interacted objects are labelled by green bounding boxes.}
\label{fig:result}
\vspace{-7pt}
\end{figure*}


%% file: appendix/fig/apx_qualitative.tex
\begin{figure}[t]
\begin{center}
\includegraphics[width=1\linewidth]{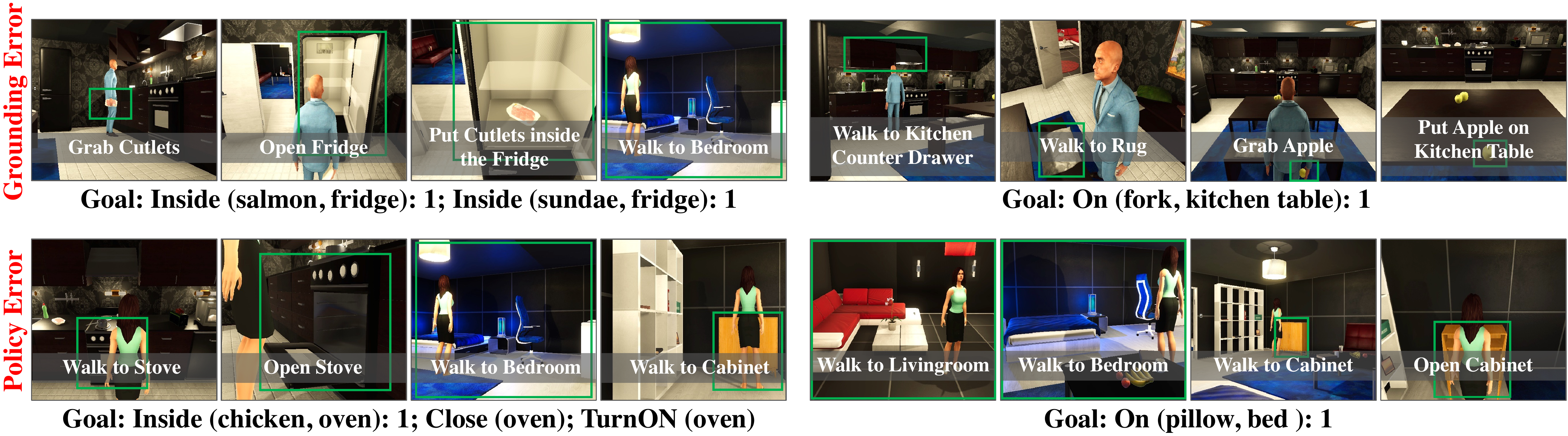}
\end{center}
\vspace{-7pt}
\caption{\small \textbf{Failure cases.} 
We show failure cases caused by the grounding error and policy error.
The interacted objects are labelled by green bounding boxes.}
\label{apx_fig:qualitative}
\vspace{-7pt}
\end{figure}

%% file: text/conclusion.tex
\section{Conclusion and Broader Impact}
\label{sec:conclusion}
In this paper, we introduced LID, a general approach to sequential decision-making that converts goals, histories, and observations into sequences and processes them using a policy initialized with a pre-trained LM. 
We integrated an active data gathering procedure into the proposed method to enable policy learning without using expert data.
Our analysis showed that input representation and favorable weight initialization both contribute to the generalization while the input encoding scheme has little influence. 
One drawback of the active data gathering is that it relies on hand-designed rules for task relabeling. 
More generally, a potential disadvantage of the proposed approach is that
biases of the pre-trained LMs may influence its behavior, and further study of LID-based models' bias is required before they may be deployed in sensitive downstream applications. 
Nevertheless, our results demonstrate that LID enables effective combinatorial generalization across different environments, and highlight the promise of LM pre-training for more general decision-making problems.



\clearpage

%% file: text/checklist.tex
\section*{Checklist}


\begin{enumerate}

\item For all authors...
\begin{enumerate}
  \item Do the main claims made in the abstract and introduction accurately reflect the paper's contributions and scope?
    \answerYes{}
  \item Did you describe the limitations of your work?
    \answerYes{} See \cref{sec:conclusion}.
  \item Did you discuss any potential negative societal impacts of your work?
    \answerYes{} See \cref{sec:conclusion}.
  \item Have you read the ethics review guidelines and ensured that your paper conforms to them?
    \answerYes{}
\end{enumerate}

\item If you are including theoretical results...
\begin{enumerate}
  \item Did you state the full set of assumptions of all theoretical results?
    \answerNA{}
        \item Did you include complete proofs of all theoretical results?
    \answerNA{}
\end{enumerate}

\item If you ran experiments...
\begin{enumerate}
  \item Did you include the code, data, and instructions needed to reproduce the main experimental results (either in the supplemental material or as a URL)?
    \answerYes{} In the supplemental material.
  \item Did you specify all the training details (e.g., data splits, hyperparameters, how they were chosen)?
    \answerYes{} See \cref{sec:experiments_main} and \cref{apx:training_details}.
        \item Did you report error bars (e.g., with respect to the random seed after running experiments multiple times)?
    \answerYes{} See \cref{sec:experiments_main}.
        \item Did you include the total amount of compute and the type of resources used (e.g., type of GPUs, internal cluster, or cloud provider)?
    \answerYes{} See \cref{apx:training_details}.
\end{enumerate}

\item If you are using existing assets (e.g., code, data, models) or curating/releasing new assets...
\begin{enumerate}
  \item If your work uses existing assets, did you cite the creators?
    \answerYes{}
  \item Did you mention the license of the assets?
    \answerNA{}
  \item Did you include any new assets either in the supplemental material or as a URL?
    \answerNA{}
  \item Did you discuss whether and how consent was obtained from people whose data you're using/curating?
    \answerNA{}
  \item Did you discuss whether the data you are using/curating contains personally identifiable information or offensive content?
    \answerYes{} See \cref{sec:conclusion}.
\end{enumerate}

\item If you used crowdsourcing or conducted research with human subjects...
\begin{enumerate}
  \item Did you include the full text of instructions given to participants and screenshots, if applicable?
    \answerNA{}
  \item Did you describe any potential participant risks, with links to Institutional Review Board (IRB) approvals, if applicable?
    \answerNA{}
  \item Did you include the estimated hourly wage paid to participants and the total amount spent on participant compensation?
    \answerNA{}
\end{enumerate}

\end{enumerate}

%% file: text/appendix.tex

\newpage
\appendix
\textbf{\huge{Appendix}}
\vspace{5pt}



In this appendix, we first show the convolutional encoding in BabyAI in \cref{apx:convolutional_encoding_babyai}.
We then describe the environment details in \cref{apx:environment} and the implementation details of the proposed model in \cref{apx:model_details}. 
We show the algorithm of interactive evaluation in \sect{apx:interactive_eval} and the data gathering procedure in \cref{apx:data_gathering}.
The goal predicates used in VirtualHome test subsets are shown in \cref{apx:test_subsets}.
We visualize the attention weights in language models in \cref{apx:attention}.

\section{Convolutional encoding in BabyAI}
\label{apx:convolutional_encoding_babyai}

In the main paper \cref{sect:input_encoding}, we explore the role of natural language by investigating two alternative ways of encoding policy inputs in VirtualHome. In this section, we show the third way of encoding policy inputs in BabyAI.

We test a new model, \textbf{\modelname-Conv (Ours)}, that converts environment inputs into \emph{convolutional embeddings}.
We pass the $7 \times 7 \times 3$ grid observation in BabyAI to convolutional layers and obtain a $7 \times 7 \times d$ feature map, where $d$ is the feature dimension. We flatten the feature map and get a sequence of features to describe the observation.
The rest of the model is the same as \emph{\modelname-Text (Ours)}.
\tbl{tbl:babyai_encodings} shows the results of policies using the \emph{text encoding} and \emph{convolutional encoding}.
\emph{\modelname-Text (Ours)} and \emph{\modelname-Conv (Ours)} have similar results given enough training data, but \emph{\modelname-Text (Ours)} is slightly better when there are fewer training data. This conclusion is coincident with the results on VirtualHome.

Different input encoding schemes have only a negligible impact on model performance: the effectiveness of pre-training is not limited to utilizing natural strings, but in fact extends to arbitrary sequential encodings.

\input{table/tokenization-babyai}

\section{Environments}
\label{apx:environment}
We use \textbf{BabyAI} \citep{hui2020babyai} and \textbf{VirtualHome} \citep{puig2018virtualhome} to evaluate the proposed method.
While both environments feature complex goals, the nature of these goals, as well as the state and action sequences that accomplish them, differ substantially across environments.

\subsection{VirtualHome}
\label{apx:virtualhome}
VirtualHome is a 3D realistic environment featuring partial observability, large action spaces, and long time horizons. It provides a set of realistic 3D homes and objects that can be manipulated to perform household organization tasks.

\textbf{Goal Space.}
For each task, we define the goal as a set of predicates and multiplicities. For example, \texttt{Inside(apple, fridge):2}; \texttt{Inside(pancake, fridge):1}; means ``put two apples and one pancake inside the fridge''.
In each task, the initial environment (including initial object locations), the goal predicates, and their orders and multiplicities are randomly sampled.
There are $59$ different types of predicates in total. 

\textbf{Observation Space.}
The observation in VirtualHome by default is a graph describing a list of objects and their relations in the current partial observation.
Each object has an object name, a state, \eg \emph{open, close, clean}, and 3D coordinates. 

\textbf{Action Space.}
Agents can navigate in the environment and interact with objects. To interact with an object, the agent must predict an action name and the index of the interested object, \eg \texttt{Open(5)} to opening the object with index (5). The agent can only interact with objects that are in the current observation or execute the navigation actions, such as \texttt{Walk(bathroom)}.
For some actions, such as \texttt{open}, the agent must be close to the object.
There are also strict preconditions for actions, \eg the agent must \texttt{grab} an object before it can \texttt{put} the object on a target position.
As a result of these constraints, the subset of actions available to the agent changes at every timestep.

We evaluate the success rates of different methods on VirtualHome.
A given episode is scored as successful if the policy completes its entire goal within $T$ steps, where $T=70$ is the maximum allowed steps of the environment.

\subsection{BabyAI}
BabyAI is a 2D grid world environment designed to evaluate instruction following. 
Different from VirtualHome, the observation in BabyAI by default is a $7 \times 7$ grid describing a partial and local egocentric view of the state of the environment. 
Each tile in the grid contains at most one object, encoded using 3 integer values: one for the object type, one for the object color, and a state for doors indicating whether it is open, closed or locked.
The goals in BabyAI are language instructions, \eg ``put the blue key next to the purple ball''.
BabyAI has 7 actions, \eg ``turn left'', ``pick up'', and ``drop''.

\input{fig/model}

\section{More implementation Details of \modelname in VirtualHome}
\label{apx:model_details}
In \cref{apx:model_architecture}, we provide more details of the model architecture used in the main paper \cref{sec:policy_network}. 
We then introduce the training detail in \cref{apx:training_details}.

\input{appendix/fig/obs-embedding}

\subsection{Model architecture details in VirtualHome}
\label{apx:model_architecture}

In this section, we provide more details of the policy network we used in VirtualHome. 
Our policy model consists of three parts, \ie inputs, the pre-trained LM, and outputs.
As shown \cref{fig:model}, we encode the inputs to the policy---including goal $g$, history $h_{t}$, and the current partial observation $o_t$---as sequences of embeddings.
These embeddings are passed to the LM (using its pre-trained embedding layer $F_{\theta}$) and used to obtain contextualized token representations. These token representations are averaged to generate a context feature $f_c$, which is then passed to fully-connected layer to predict the next action $a_{t}$.
The output action in VirtualHome consists of a verb and an object. 
For brevity, we omit the time subscript $t$ from now on.

In VirtualHome, the partial observation $o$ of the environment state can be represented as a list of objects in the agent's view.
We represent each object by its name, \eg ``oven'', a state description, \eg ``open, clean'', and position both in the world and relative to the agent.
In this part, we provide more details of how \textbf{\modelname-Text (Ours)} encodes the name, state, and position of each object in the observation. 
\fig{apx_fig:obs_encoding} shows the model architecture we used to encode the observation.

\textbf{Name encoding.}
For each object node, we serialize its object name as an English phrase $s^o$.
We extract its tokens and features using the tokenizer and the embedding layer of the pre-trained LM, respectively.
Since one object name might generate several English tokens using the tokenizer from the pre-trained LM, \eg the tokens of ``kitchencabinet'' is $[15813, 6607, 16212, 500]$, we take the averaged features of all the tokens in the object name and obtain a ``name'' feature $f^{o,\text{name}}_{i}$ for each object node as shown in \fig{apx_fig:obs_encoding}.

\textbf{State encoding.}
Some objects have a state description, \eg ``oven: open, clean''.
There are six types of object states in the environment: ``clean'', ``closed'', ``off'', ``on'', ``open'', and ``none''.
For each object node, we use a binary vector to represent its state.
Taking the ``oven'' as an example, if the oven is open and clean, its state vector would be $[1,0,0,0,1,0]$.
This state vector is then passed through a fully-connected layer to generate a state feature $f^{o,\text{state}}_{i}$ of object $o_i$.

\textbf{Position encoding.}
To encode the position information of each object $o_i$, we take their world coordinates $\{o_{i,x}, o_{i,y}, o_{i,z}\}$ and their spatial distance to the agent $\{a_{x}, a_{y}, a_{z}\}$ to generate a position vector $[o_{i,x}, o_{i,y}, o_{i,z}, o_{i,x}-a_{x}, o_{i,y}-a_{y}, o_{i,z}-a_{z}]$.
This position vector is then passed through two fully-connected layers with a ReLU layer in the middle to generate a position feature $f^{o,\text{position}}_{i}$ of object $o_i$.

The final feature $f^o_i$ of each object node is obtained by passing the concatenation of its name feature $f^{o,\text{name}}_{i}$, state feature $f^{o,\text{state}}_{i}$, and position feature $f^{o,\text{position}}_{i}$ through a fully connected layer.
The observation at a single step can be written as a set of features $\{f_1^o, \cdots, f_N^o\}$, where $N$ is the number of objects in the current observation.

\subsection{Training details}
\label{apx:training_details}
Our proposed approach and baselines are trained on Tesla 32GB GPUs. We train every single model on 1 Tesla 32GB GPU.
All experiments used the AdamW optimizer with the learning rate of $10^{-5}$.
We utilize a standard pre-trained language model, GPT-2, in our experiments. GPT-2 is trained on the Webtext dataset \citep{radford2019language} using the Huggingface library \citep{ wolf2019huggingface}.

\section{Interactive Evaluation}
\label{apx:interactive_eval}

The algorithm for interactive evaluation is shown in \cref{alg:eval}.

\input{algo/eval}

\section{Data Gathering Details in VirtualHome}
\label{apx:data_gathering}

In this section, we provide more data gathering details in VirtualHome for training the decision-making policies.
We introduce the expert data collection and active data gathering in \cref{apx:expert_data_collection} and \cref{apx:active_data_gathering}, respectively.

\input{appendix/fig/regression_planner}

\subsection{Expert Data Collection}
\label{apx:expert_data_collection}

\textbf{VirtualHome-Imitation Learning Dataset.}
To train the models, we collect a set of expert trajectories in VirtualHome using regression planning (RP) \citep{korf1987planning}.
We follow the implementation of the regression planner used in \citep{puig2020watch}.
Given a task described by goal predicates, the planner generates an action sequence to accomplish this task.
As shown in \fig{apx_fig:regression_planner}, the agent has a belief about the environment, \ie an imagined distribution of object locations. As the agent explores the environment, its belief of the world becomes closer to the real world.
At every step, the agent updates its belief based on the latest observation (see \citep{puig2020watch}), finds a new plan using the regression planner, and executes the first action of the plan.
If the subtask (described by the goal predicate) has been finished, the agent will select a new unfinished subtask, otherwise, the agent will keep doing this subtask until it finishes.

Similarly to previous work \citep{shridhar2020alfred,shen2020igibson,puig2020watch}, we generate training data using a planner that has access to privileged information, such as full observation of the environment and information about the pre-conditions and effects of each action. 
The planner allows an agent to robustly perform tasks in partially observable environments and generate expert trajectories for training and evaluation.
We generate $20,000$ trajectories for training and $3,000$ trajectories for validation.
Each trajectory has a goal, an action sequence, and the corresponding observations after executing each action.

\subsection{Active Data Gathering}
\label{apx:active_data_gathering}
\vspace{-3pt}
The algorithm for active data gathering is shown in \cref{alg:training}.
To sample the goal and initial state, we first generate a set of initial states in VirtualHome using the code released by~\citep{puig2020watch}. For each initial state, we are able to get a set of feasible tasks that can be accomplished in this environment. For example, in an initial state, if the apple is on the kitchen table, a feasible task goal could be ``put the apple inside the fridge''. In contrast, ``put the banana inside the fridge'' is not a feasible task if there is no banana in the initial state.

We collect 9893 initial states, and randomly sample an initial state and its feasible goal every time when we reset the environment. After each data collection iteration, we obtain a set of new goals using the goal relabel function. We save the goal and its corresponding initial state in the replay buffers and use the same strategy to sample the goal and initial state in the next iteration.

The \emph{hindsight relabeling} stage is the key component for active data gathering. Here we provide more implementation details of how we relabel ``failed'' trajectories with new goals in the \emph{hindsight relabeling} stage.
For each ``failed'' trajectory, we extract its useful sub-trajectories and relabel a task goal $g'$ for it.
We design a goal relabel function $f_l$ that generates a goal based on the sequence of observations and actions.
To do this, we first use a hand-designed program to detect what tasks are contained in a ``failed'' trajectory. This program find useful tasks based on the keywords in the action list.
For example in \cref{apx_fig:trajectory}, the program knows the trajectory containing a task of ``\texttt{On(apple, kitchen table):1}'' based on the action ``$[put] <apple> <kitchentable>$''.

The selected sub-trajectories are not always optimal. We thus design a rule to filter out bad trajectories, \ie for trajectories with the same goal, selecting the ``shorter'' ones.
One example is shown in \cref{apx_fig:filter}.
Suppose that there are two trajectories having the same goal, \eg ``\texttt{On(apple, kitchen table):1}''.
The first trajectory has actions that are redundant or not related to the task, such as ``$[walk] <bathroom>$'' and ``$[walk] <kitchen>$'' while the second trajectory is more optimal given the goal.
We select the second trajectory and delete the first trajectory from the replay buffer. 
Note that the ``shorter'' does not mean fewer actions, but fewer actions that are not related to the task.
The \emph{hindsight relabeling} stage allows sample-efficient learning by reusing the failure cases.
The relabeled data are used to train policies in the \emph{policy update} stage.

\input{algo/training}

\input{appendix/fig/apx_trajectory}

\input{appendix/fig/apx_filter}

\vspace{-5pt}
\section{Test Sets in VirtualHome}
\label{apx:test_subsets}
\vspace{-3pt}
In this section, we provide more details of each test set. We first introduce the test sets used for evaluating the proposed model trained on expert data, \ie \modelname, in \sect{apx:test_set_imitation}. We then show the test sets used for evaluating the proposed model with active data gathering, \ie \modelname-ADG, in \sect{apx:test_set_active}.

\subsection{\modelname Test Sets}
\label{apx:test_set_imitation}

In \sect{sec:exp_offline_data}, we compared the proposed method and baselines trained on expert data.
In \tbl{tbl:test_subsets}, we provide a detailed description of each test subset, including the count of goal predicate types and the number of goal predicates in each task.
The \textbf{In-Distribution} setting has 37 goal predicates in total and each task has $2\sim10$ goal predicates. The tasks are drawn from the same distribution as the training tasks.
The \textbf{Novel Scenes} setting also has 37 goal predicates and each task has $2\sim10$ goal predicates. The objects are randomly placed in the initial environment.
The \textbf{Novel Tasks} setting has 22 goal predicates in total and each task has $2\sim8$ goal predicates. The tasks are never seen during training.

\vspace{-5pt}
\subsection{\modelname-ADG Test Sets}
\label{apx:test_set_active}

As we have mentioned in the main paper \sect{sec:conclusion}, one limitation of active data gathering is that it relies on
hand-designed rules for task relabeling.
In addition, it is sometimes challenging to define effective rules to extract useful sub-trajectories and get high-quality hindsight labels, especially when trajectories are long and tasks become more complex.
Thus we only relabel short sub-trajectories, where the goal consists of a single goal predicate, \eg ``\texttt{On(apple, kitchen table):1}''.
During testing, we evaluate the success rate of approaches on solving such tasks as well, \ie the count of the goal predicate equals to 1. The types of goal predicates are the same as \sect{apx:test_set_imitation}, \ie 37 goal predicates in the \emph{In-Distribution} setting and the \emph{Novel Scenes} setting, and 22 goal predicates in the \emph{Novel Tasks} setting.

\input{appendix/table/tasks_test}

\vspace{-5pt}
\section{Visualization of Attention Weights}
\label{apx:attention}
To better understand how does LM pre-trained policies make decisions, we visualize the attention weights from the self-attention layers of GPT-2 \citep{vaswani2017attention} in \fig{apx_fig:attention1} and \fig{apx_fig:attention2}.
In the inference time, when we are decoding the actions, we save the self-attention weights with respect to different layers and different heads. Then, we use BertViz library~\cite{vig-2019-multiscale} to visualize normalized attention weights.
We show the attention weights from the input to the output of \textbf{\modelname-Text (Ours)}. The order of tokens in the input and ouput is observation, goal, and history.
In \fig{apx_fig:attention1} and \fig{apx_fig:attention2}, the left side is the query side. The boldness of the lines is proportional with the attention weight.

\fig{apx_fig:attention1} illustrates the attention weights of a layer named ``Head 3 Layer 2''. 
We show attention weights on two different tasks.
We find that ``Head 3 Layer 2'' can capture objects in the goal predicates, such as ``wineglass'' and ``cutleryfork'' in the left figure, and ``pancake'' and ``chicken'' in the right figure (the figures are cropped for visualization).

\fig{apx_fig:attention2} illustrates the attention weights of layers named ``Head 1 Layer 2'' (left) and ``Head 4 Layer 11'' (right). 
Given the goal predicates, history, and the current observation, the policy predicts the next action as ``grab milk''.
We find that ``Head 1 Layer 2'' is able to capture objects in the goal predicates, such as ``milk'', ``pancake'', and ``chicken'' while ``Head 4 Layer 11'' focuses on the interacted object in the predicted action, such as ``milk''.

The attention weights from different self-attention layers are significantly different---some self-attention layers assign high attention weight to objects in the goal predicates while some layers focus on the interacted object. There are also some layers that do not have interpretable meanings.
The attention weights just provide us an intuition of how does the internal language model works, more quantified results are reported in the main paper.

\input{appendix/fig/attention}

%% file: table/tokenization-babyai.tex
\begin{table}[h]
\vspace{-5pt}
  \caption{\small 
         \textbf{Success rate of policies trained with \emph{text encoding} vs. \emph{convolutional encoding} on BabyAI.} The text encoding is more sample-efficient, but both models converge to near perfect performance given sufficient training data.
         }
\label{tbl:babyai_encodings}
\small
  \centering
  \addtolength{\tabcolsep}{-34pt}
  \scalebox{0.9}{
  \setlength{\tabcolsep}{0.4em}
  \begin{tabular}{llcccccc}
    \toprule
    \bf Tasks & \bf Methods & \multicolumn{5}{c}{\bf Number of Demos} \\
     \cmidrule(lr){3-7}
    & &  \bf 100 & \bf 500 & \bf 1K & \bf 5K & \bf 10K \\
    
    \midrule
    
    \multirow{2}{*}{GoToRedBall} 
    &  \modelname-Text (Ours) & \bf 93.9 & \bf 99.4 & 99.7 & \bf 100.0 & \bf 100.0 \\
    & \modelname-Conv (Ours) & 92.5 & 98.8 & \bf 100.0 & \bf 100.0 & \bf 100.0 \\ 

    \midrule
    
    \multirow{2}{*}{GoToLocal} 
    & \modelname-Text (Ours) &  64.6 & \bf 97.9 & \bf 99.0 & 99.5 & 99.5 \\
    & \modelname-Conv (Ours) & \bf 69.5 & 86.0 & 98.2 & \bf 99.9 & \bf 99.9 \\ 

    \midrule
    
    \multirow{2}{*}{PickupLoc} 
    & \modelname-Text (Ours) & \bf 28.7 & \bf 73.4 & \bf 99.0 & \bf 99.6 & 99.8 \\
    & \modelname-Conv (Ours) & 25.0 & 58.8 & 95.1 & \bf 99.6 & \bf 100.0 \\ 

    \midrule
    
    \multirow{2}{*}{PutNextLocal} 
    & \modelname-Text (Ours) & 11.1 & \bf 93.0 & \bf 93.2 & \bf 98.9 & \bf 99.9 \\
    & \modelname-Conv (Ours) & \bf 17.9 & 53.6 & 91.3 & 97.7 & 99.5 \\ 

    \bottomrule
  \end{tabular}
  }
\vspace{-5pt}
\end{table}

%% file: fig/model.tex
\begin{figure}
\begin{center}
\includegraphics[width=0.7\linewidth]{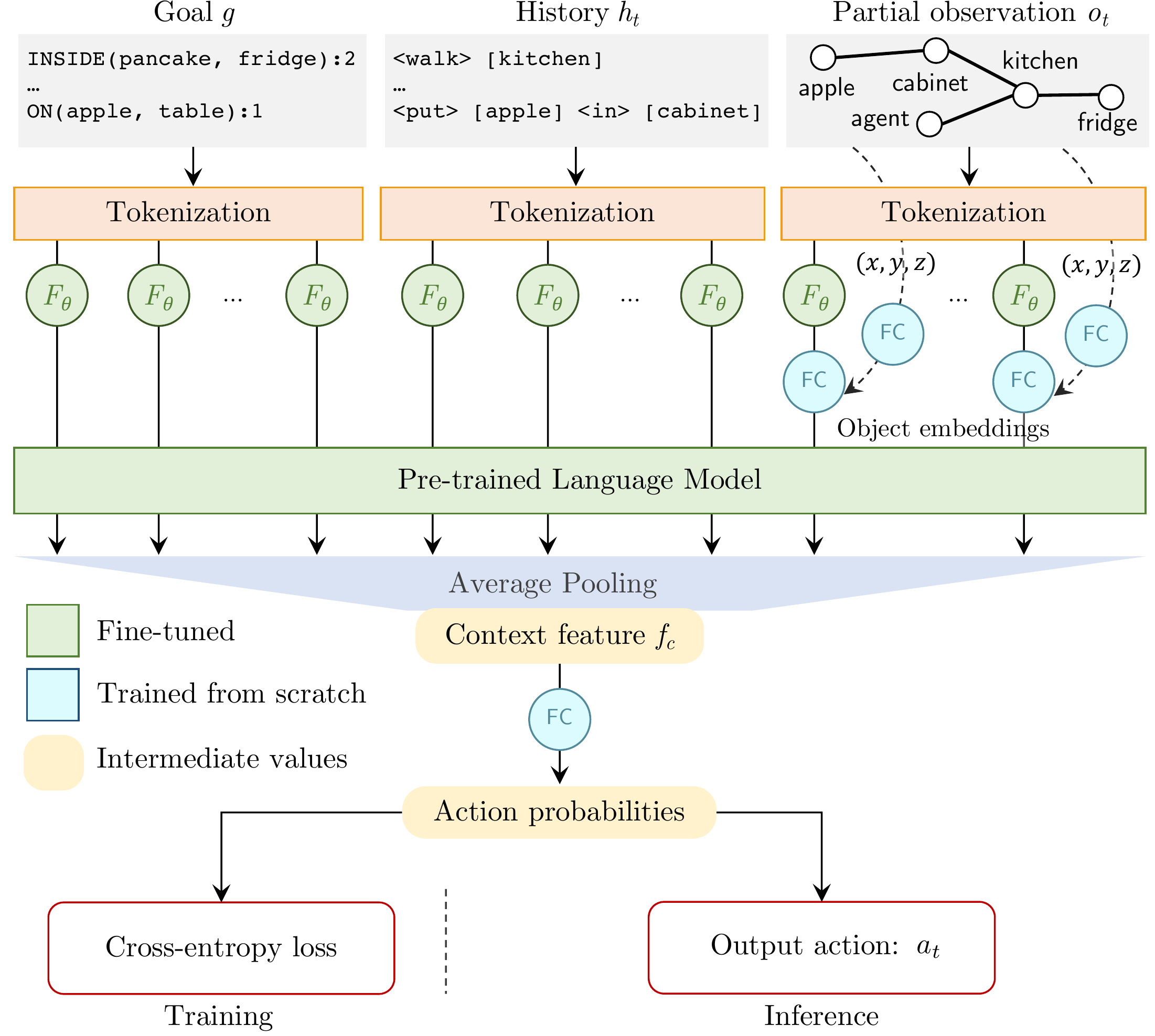}
\end{center}
\vspace{-5pt}
\caption{\small \textbf{Policy network in VirtualHome.} The observation, goal, and history are first converted into sequences and then passed through an embedding layer $F_\theta$. 
The combined sequence is passed through a pre-trained LM, and the output tokens are pooled into a context feature vector for action prediction. 
}
\label{fig:model}
\vspace{-5pt}
\end{figure}

%% file: appendix/fig/obs-embedding.tex
\begin{figure}
\begin{center}
\includegraphics[width=0.8\linewidth]{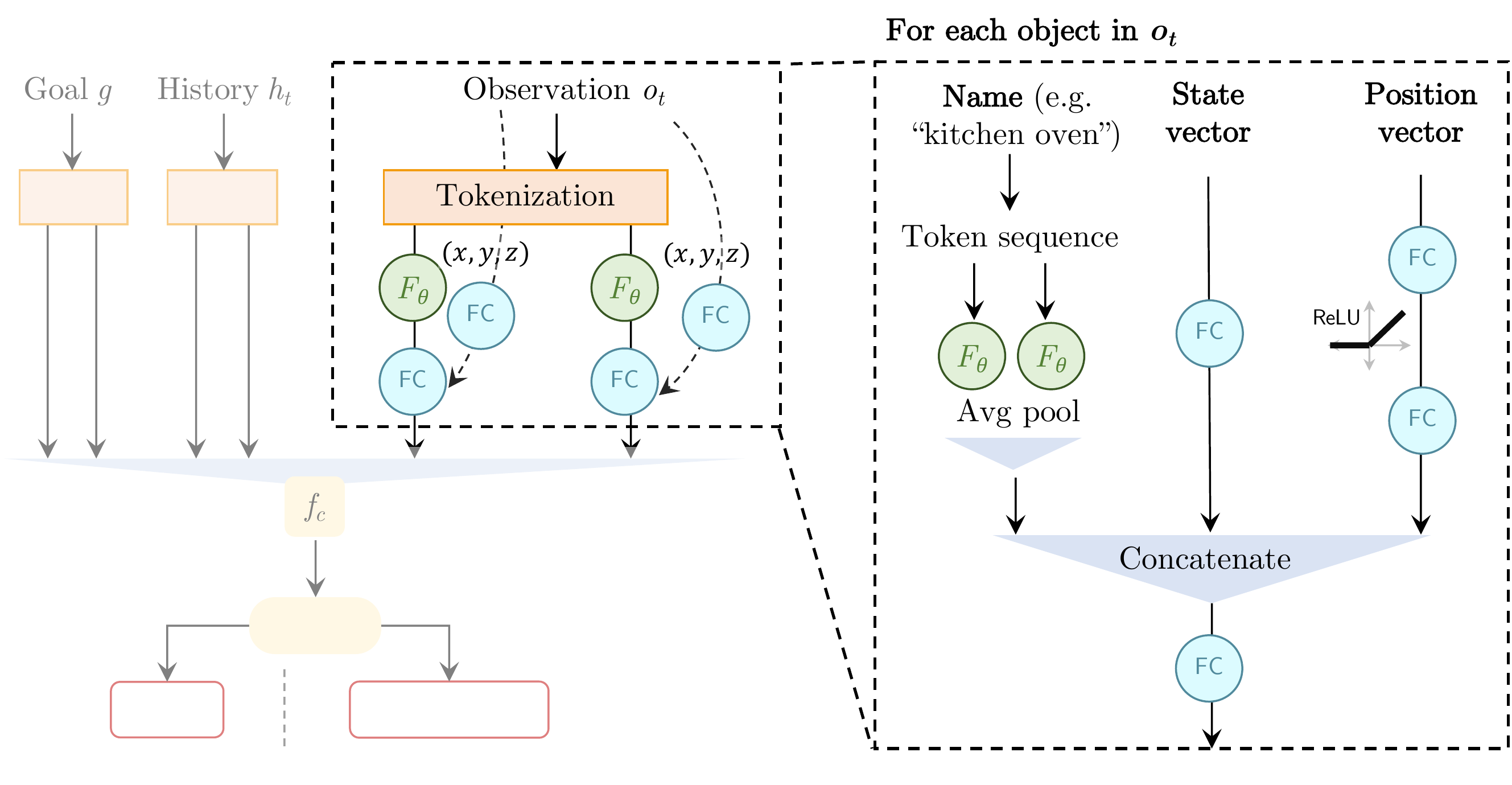}
\end{center}
\vspace{-10pt}
\caption{\small \textbf{Object encoding.} In VirtualHome, the partial observation of the environment state can be represented as a list of objects in the agent's view. Each object is represented by a name, a state vector, and position vector. 
\textbf{Object name encoding:} each object's name is an English phrase.
We tokenize the phrase, embed the tokens, and average the embeddings. 
\textbf{Object state encoding:}
each object is assigned one of six states: ``clean'', ``closed'', ``off'', ``on'', ``open'', or ``none''.
This state is represented as a 6-dimensional binary vector and passed through a fully-connected layer. 
\textbf{Object position encoding:} 
an object's position vector is a 6-dimensional vector containing its world coordinates alongside its displacement to the agent (\ie the difference in their world coordinates). 
This position vector is passed through two fully-connected layers. 
These three features are concatenated and passed through a fully-connected layer to obtain the representation of an object in the current observation.
}
\label{apx_fig:obs_encoding}
\vspace{-10pt}
\end{figure}

%% file: algo/eval.tex
\begin{algorithm}[h]
\small
\caption{Interactive evaluation}
\label{alg:eval}
\SetAlgoLined

A set of task goals $G$ (each goal has a corresponding initial state);

Load the learned policy $\pi_{\phi}$;

Successful trajectory count: $n = 0$;

\For{\textit{example=1, $N_{\text{test}}$}}{

    Sample a goal $g$ and the an initial state;
    
    \For{$t=0, T$}
    {
        Sample an action $a_{t}$ from policy $\pi_{\phi}(a_t|g, h_t, o_t)$;
            
        Execute the action $a_{t}$ and get a new observation $o_{t+1}$;
        
        \If{success}{
            $n = n + 1$;
            
            break;
        }
    }
    }
    success rate: $r=n/N_{\text{test}}$;
\end{algorithm}

%% file: appendix/fig/regression_planner.tex
\begin{figure}[h]
\begin{center}
\includegraphics[width=0.85\linewidth]{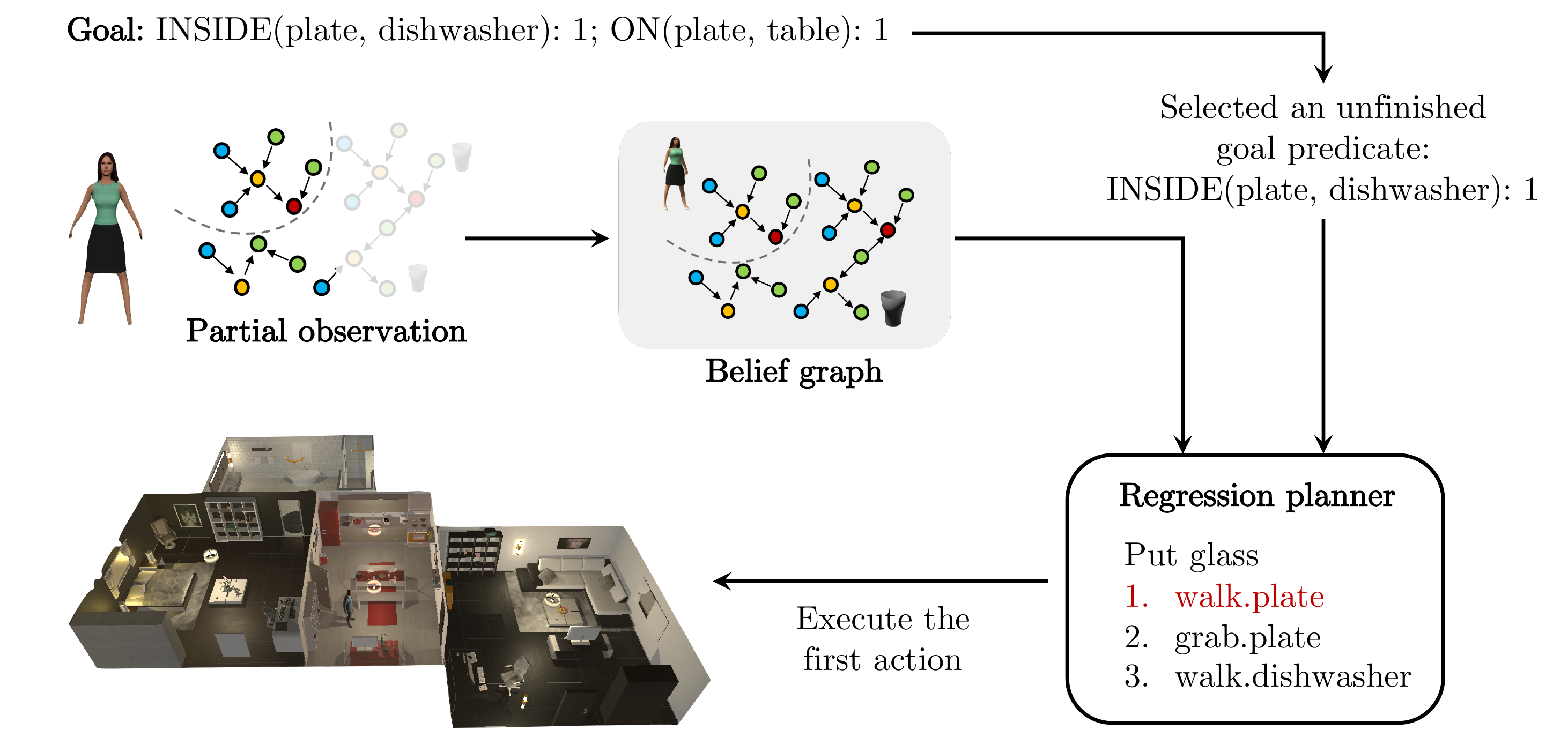}
\end{center}
\vspace{-10pt}
\caption{\small \textbf{Regression planner}.
Given a task described by goal predicates, the planner generates an action sequence to accomplish this task.
The agent has a belief about the environment, \ie an imagined distribution of object locations. As the agent explores the environment, its belief of the world becomes closer to the real world.
At every step, the agent updates its belief based on the latest observation, finds a new plan using the regression planner, and executes the first action of the plan.
If the subtask (described by the goal predicate) has been finished, the agent will select a new unfinished subtask, otherwise, the agent will keep doing this subtask until finish it. 
}
\label{apx_fig:regression_planner}
\end{figure}

%% file: algo/training.tex
\begin{center}
\begin{algorithm}[t]
\small
\caption{Active Data Gathering}
\label{alg:training}
\SetAlgoLined
\textbf{Given:}
a goal relabel function $f_l$;

\textbf{Initialize:}
policy $\pi_{\phi}$;
goal set $G$;
training replay buffer $\mathcal{R}_{train}=\{\}$;
validation replay buffer $\mathcal{R}_{val}=\{\}$;

\For{\textit{iteration=1, $N$}}{
    \For{\textit{example=1, $M$}}{
        Sample a goal $g$ from $G$ and an initial state $s_1$;
        
        \For{$t=1, T$}
        {
            Sample an action from policy $\pi_{\phi}(a_{t}| g, h_t, o_t)$ or sample an action randomly; 
            
            Execute $a_{t}$ and obtain a new observation $o_{t+1}$;
        }
        Store the trajectory $(o_1,a_1,\cdots,o_T,a_T,g)$ in the replay buffer $\mathcal{R}_{train}$ or $\mathcal{R}_{val}$; 
    }
    
    
    Relabel each failure trajectory $d=(o_1,a_1,\cdots,o_T,a_T)$ in the replay buffers and get new goal $g'=f_l(d)$;
    
    
    
    Put new goals $g'$ in the goal set $G$;
    
    
    \For{ $k=1,K$ }{
    
        \Repeat{\textit{training episode ends}}{
            Sample data from $\mathcal{R}_{train}$ and
            update policy $\pi_{\phi}$;
        }
        
        Get validation accuracy using the data from $\mathcal{R}_{val}$;
        
        
        
            
        
    }
    
    
    $\pi_{\phi} = \pi_{\text{val\_best}}$
    
}
\end{algorithm}
\end{center}


%% file: appendix/fig/apx_trajectory.tex
\begin{figure}[t]
\begin{center}
\includegraphics[width=1\linewidth]{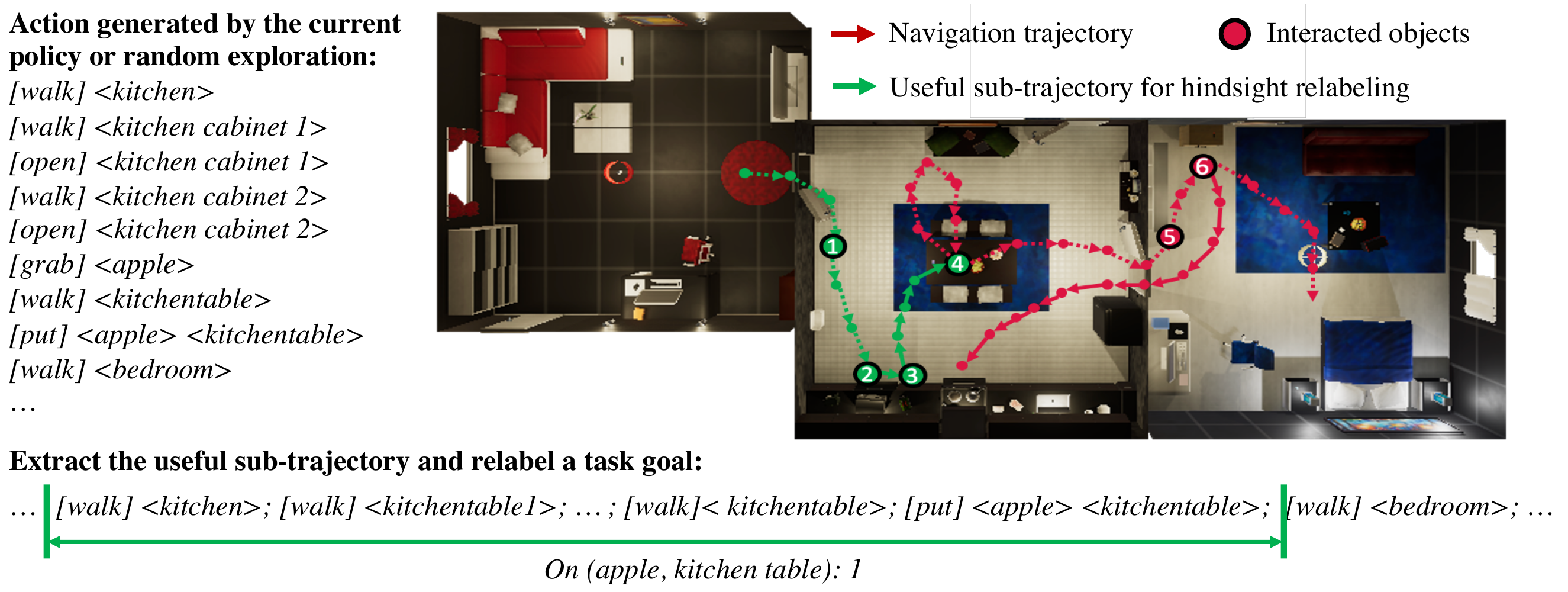}
\end{center}
\vspace{-5pt}
\caption{\small We first use a hand-designed program to detect what tasks are contained in the collected trajectory. This program find tasks based on the keywords in the action list.
For example, the program knows the trajectory containing a task of ``\texttt{On(apple, kitchen table):1}'' based on the action ``$[put] <apple> <kitchentable>$''.
Then the program extracts all previous actions related to this task using hand-designed rules.}
\label{apx_fig:trajectory}
\vspace{-5pt}
\end{figure}

%% file: appendix/fig/apx_filter.tex
\begin{figure}[t]
\begin{center}
\includegraphics[width=1\linewidth]{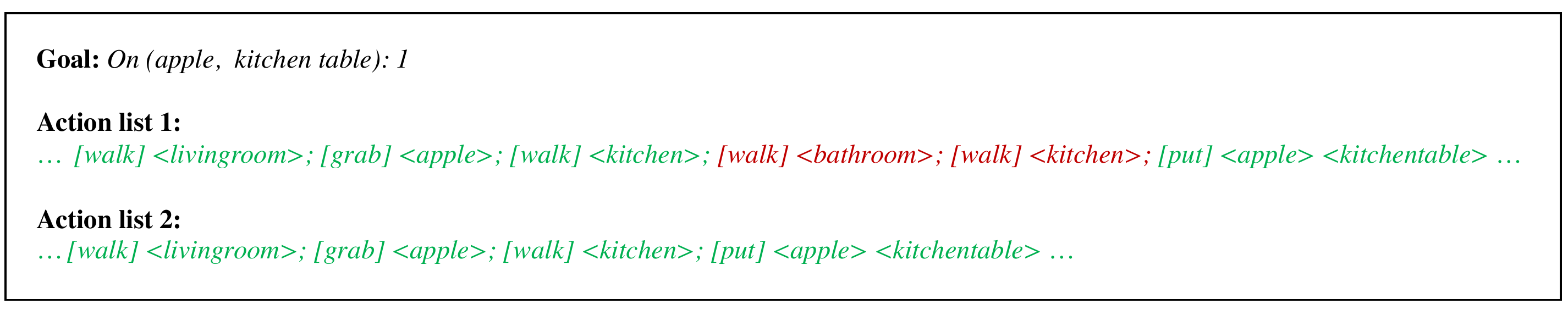}
\end{center}
\vspace{-5pt}
\caption{\small Suppose there are two trajectories having the same goal, \eg ``\texttt{On(apple, kitchen table):1}''.
The first trajectory has actions that are redundant or not related to the task, such as ``$[walk] <bathroom>$'' and ``$[walk] <kitchen>$'' while the second trajectory is more optimal given the goal.
We select the second trajectory and delete the first trajectory from the replay buffer.
Note that the ``shorter'' does not mean fewer actions, but fewer actions that are not related to the task.}
\label{apx_fig:filter}
\vspace{-5pt}
\end{figure}

%% file: appendix/table/tasks_test.tex
\begin{table}[t]
\caption{\small{\textbf{Test sets used for
evaluating the proposed model trained on the expert data.} We show the count of goal predicate types and the number of goal predicates used in each task.}}
\label{tbl:test_subsets}
\begin{center}
\small
\setlength{\tabcolsep}{3pt}
\resizebox{1\textwidth}{!}{
\begin{tabular}{l|c|c|c}
\toprule
\bf Test Sets & \bf Predicate Types & \bf \#Predicate Per Task & \bf Compared with the training set \\
\midrule
\textbf{In-Distribution} & 37 & $2\sim10$ & Tasks are drawn from the same distribution as training tasks. \\
\midrule
\textbf{Novel Scenes} & 37 & $2\sim10$ &  The objects are randomly placed in the initial environment.  \\
\midrule
\textbf{Novel Tasks} & 22 & $2\sim8$  & Tasks are never seen during training. \\
\bottomrule
\end{tabular}
}
\end{center}
\end{table}

%% file: appendix/fig/attention.tex
\begin{figure}[h]
\begin{center}
\includegraphics[width=0.8\linewidth]{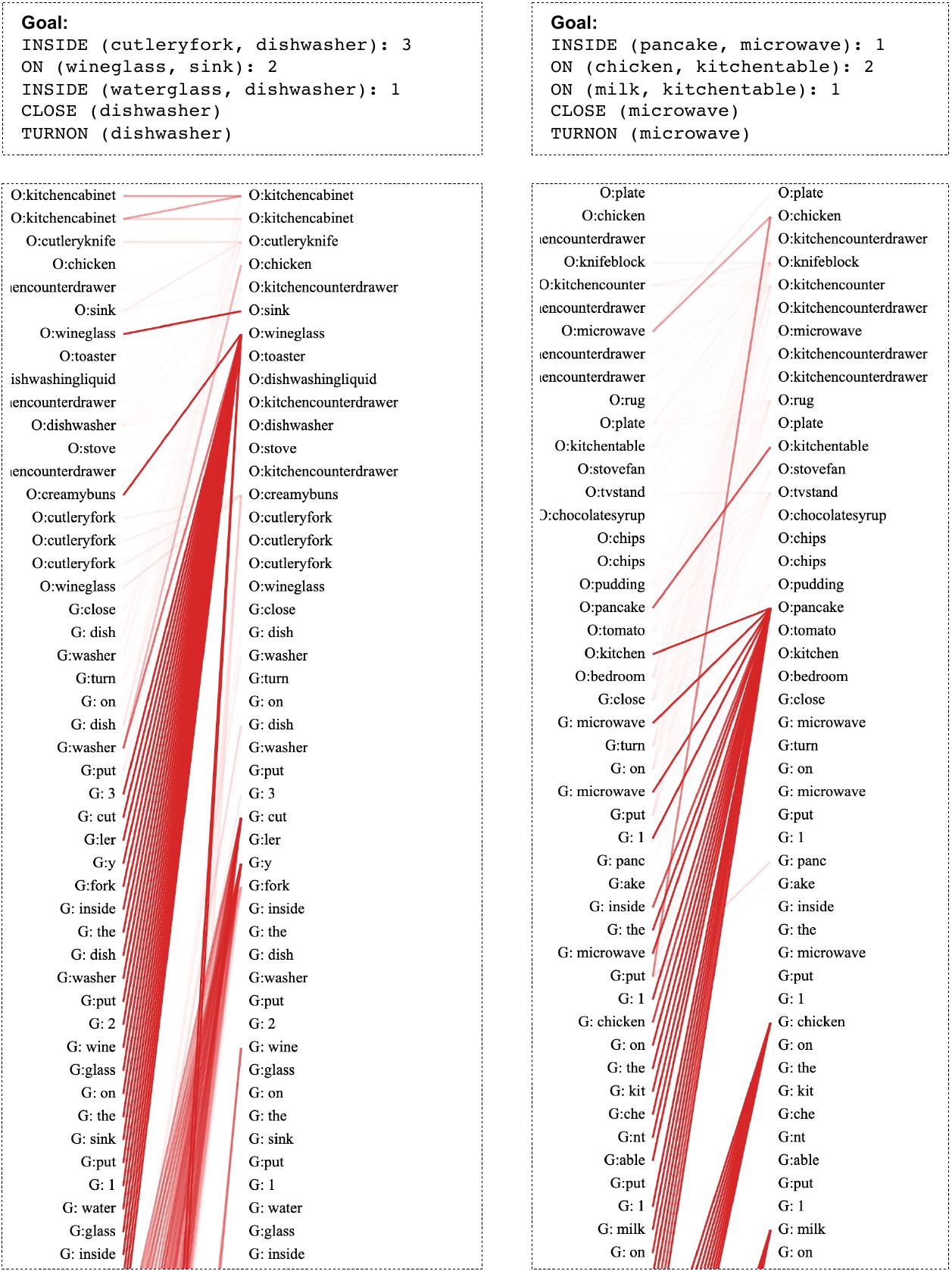}
\end{center}
\vspace{-10pt}
\caption{\small \textbf{Attention weights of a layer named ``Head 3 Layer 2''.} 
We show attention weights on two different tasks.
We find that ``Head 3 Layer 2'' is able to capture objects in the goal predicates, such as ``wineglass'' and ``cutleryfork'' in the left figure, and ``pancake'' and ``chicken'' in the right figure (the figures are cropped for visualization).}
\label{apx_fig:attention1}
\vspace{-10pt}
\end{figure}

\begin{figure}
\begin{center}
\includegraphics[width=0.8\linewidth]{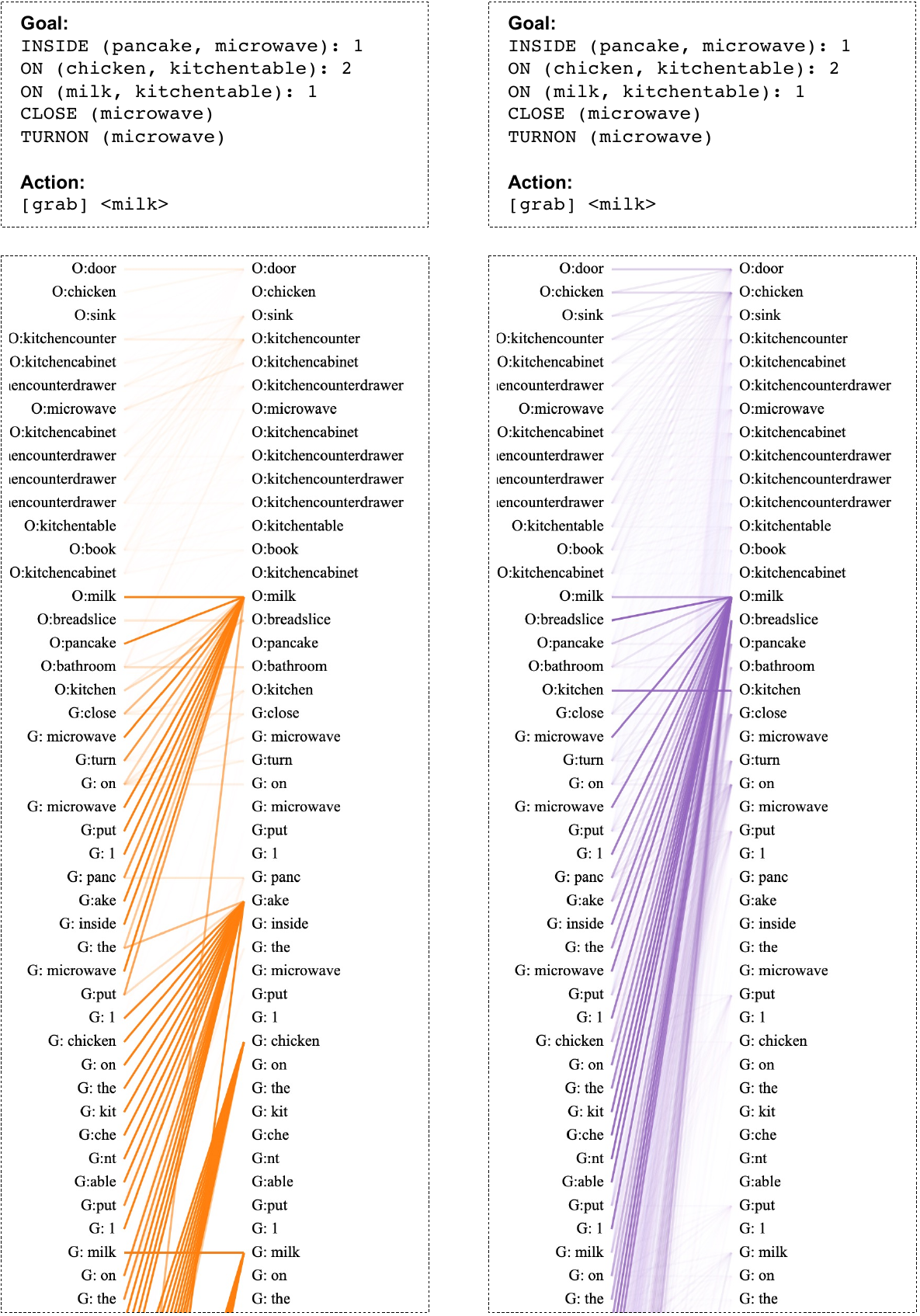}
\end{center}
\vspace{-10pt}
\caption{\small \textbf{Attention weights of layers named ``Head 1 Layer 2'' (left) and ``Head 4 Layer 11'' (right).} 
Given the goal predicates, history, and the current observation, the policy model predicts the next action as ``grab milk''.
We find that ``Head 1 Layer 2'' can capture objects in the goal predicates, such as ``milk'', ``pancake'', and ``chicken'' while ``Head 4 Layer 11'' focuses on the interacted object in the predicted action, such as ``milk''.}
\label{apx_fig:attention2}
\vspace{-10pt}
\end{figure}

%% file: neurips_2021.bbl
\begin{thebibliography}{10}

\bibitem{ammanabrolu2018playing}
P.~Ammanabrolu and M.~O. Riedl.
\newblock Playing text-adventure games with graph-based deep reinforcement
  learning.
\newblock {\em arXiv preprint arXiv:1812.01628}, 2018.

\bibitem{andreas19l3}
J.~Andreas and D.~Klein.
\newblock Learning with latent language.
\newblock In {\em North American Association for Computational Linguistics},
  2022.

\bibitem{andrychowicz2017hindsight}
M.~Andrychowicz, F.~Wolski, A.~Ray, J.~Schneider, R.~Fong, P.~Welinder,
  B.~McGrew, J.~Tobin, P.~Abbeel, and W.~Zaremba.
\newblock Hindsight experience replay.
\newblock {\em arXiv preprint arXiv:1707.01495}, 2017.

\bibitem{jacob21llp}
M.~L. Athul Paul~Jacob and J.~Andreas.
\newblock Multitasking inhibits semantic drift.
\newblock In {\em North American Association for Computational Linguistics},
  2021.

\bibitem{1606.01540}
G.~Brockman, V.~Cheung, L.~Pettersson, J.~Schneider, J.~Schulman, J.~Tang, and
  W.~Zaremba.
\newblock Openai gym, 2016.

\bibitem{brown2020language}
T.~B. Brown, B.~Mann, N.~Ryder, M.~Subbiah, J.~Kaplan, P.~Dhariwal,
  A.~Neelakantan, P.~Shyam, G.~Sastry, A.~Askell, et~al.
\newblock Language models are few-shot learners.
\newblock {\em arXiv preprint arXiv:2005.14165}, 2020.

\bibitem{chen2021decision}
L.~Chen, K.~Lu, A.~Rajeswaran, K.~Lee, A.~Grover, M.~Laskin, P.~Abbeel,
  A.~Srinivas, and I.~Mordatch.
\newblock Decision transformer: Reinforcement learning via sequence modeling.
\newblock {\em arXiv preprint arXiv:2106.01345}, 2021.

\bibitem{deerwester1990indexing}
S.~Deerwester, S.~T. Dumais, G.~W. Furnas, T.~K. Landauer, and R.~Harshman.
\newblock Indexing by latent semantic analysis.
\newblock {\em Journal of the American society for information science},
  41(6):391--407, 1990.

\bibitem{devlin2018bert}
J.~Devlin, M.-W. Chang, K.~Lee, and K.~Toutanova.
\newblock Bert: Pre-training of deep bidirectional transformers for language
  understanding.
\newblock {\em arXiv preprint arXiv:1810.04805}, 2018.

\bibitem{dosovitskiy2020image}
A.~Dosovitskiy, L.~Beyer, A.~Kolesnikov, D.~Weissenborn, X.~Zhai,
  T.~Unterthiner, M.~Dehghani, M.~Minderer, G.~Heigold, S.~Gelly, et~al.
\newblock An image is worth 16x16 words: Transformers for image recognition at
  scale.
\newblock {\em arXiv preprint arXiv:2010.11929}, 2020.

\bibitem{dumais2004latent}
S.~T. Dumais.
\newblock Latent semantic analysis.
\newblock {\em Annual review of information science and technology},
  38(1):188--230, 2004.

\bibitem{ghosh2019learning}
D.~Ghosh, A.~Gupta, A.~Reddy, J.~Fu, C.~Devin, B.~Eysenbach, and S.~Levine.
\newblock Learning to reach goals via iterated supervised learning.
\newblock {\em arXiv preprint arXiv:1912.06088}, 2019.

\bibitem{hill2020human}
F.~Hill, S.~Mokra, N.~Wong, and T.~Harley.
\newblock Human instruction-following with deep reinforcement learning via
  transfer-learning from text.
\newblock {\em arXiv preprint arXiv:2005.09382}, 2020.

\bibitem{hochreiter1997long}
S.~Hochreiter and J.~Schmidhuber.
\newblock Long short-term memory.
\newblock {\em Neural computation}, 9(8):1735--1780, 1997.

\bibitem{huang2022language}
W.~Huang, P.~Abbeel, D.~Pathak, and I.~Mordatch.
\newblock Language models as zero-shot planners: Extracting actionable
  knowledge for embodied agents.
\newblock {\em arXiv preprint arXiv:2201.07207}, 2022.

\bibitem{hui2020babyai}
D.~Y.-T. Hui, M.~Chevalier-Boisvert, D.~Bahdanau, and Y.~Bengio.
\newblock Babyai 1.1, 2020.

\bibitem{jang2022bc}
E.~Jang, A.~Irpan, M.~Khansari, D.~Kappler, F.~Ebert, C.~Lynch, S.~Levine, and
  C.~Finn.
\newblock Bc-z: Zero-shot task generalization with robotic imitation learning.
\newblock In {\em Conference on Robot Learning}, pages 991--1002. PMLR, 2022.

\bibitem{keskar2019ctrl}
N.~S. Keskar, B.~McCann, L.~R. Varshney, C.~Xiong, and R.~Socher.
\newblock Ctrl: A conditional transformer language model for controllable
  generation.
\newblock {\em arXiv preprint arXiv:1909.05858}, 2019.

\bibitem{kitaev2018multilingual}
N.~Kitaev, S.~Cao, and D.~Klein.
\newblock Multilingual constituency parsing with self-attention and
  pre-training.
\newblock {\em arXiv preprint arXiv:1812.11760}, 2018.

\bibitem{korf1987planning}
R.~E. Korf.
\newblock Planning as search: A quantitative approach.
\newblock {\em Artificial intelligence}, 33(1):65--88, 1987.

\bibitem{lillicrap2015continuous}
T.~P. Lillicrap, J.~J. Hunt, A.~Pritzel, N.~Heess, T.~Erez, Y.~Tassa,
  D.~Silver, and D.~Wierstra.
\newblock Continuous control with deep reinforcement learning.
\newblock {\em arXiv preprint arXiv:1509.02971}, 2015.

\bibitem{lu2019vilbert}
J.~Lu, D.~Batra, D.~Parikh, and S.~Lee.
\newblock Vilbert: Pretraining task-agnostic visiolinguistic representations
  for vision-and-language tasks.
\newblock {\em arXiv preprint arXiv:1908.02265}, 2019.

\bibitem{lu2021pretrained}
K.~Lu, A.~Grover, P.~Abbeel, and I.~Mordatch.
\newblock Pretrained transformers as universal computation engines.
\newblock {\em arXiv preprint arXiv:2103.05247}, 2021.

\bibitem{majumdar2020improving}
A.~Majumdar, A.~Shrivastava, S.~Lee, P.~Anderson, D.~Parikh, and D.~Batra.
\newblock Improving vision-and-language navigation with image-text pairs from
  the web.
\newblock In {\em European Conference on Computer Vision}, pages 259--274.
  Springer, 2020.

\bibitem{mikolov2013distributed}
T.~Mikolov, I.~Sutskever, K.~Chen, G.~S. Corrado, and J.~Dean.
\newblock Distributed representations of words and phrases and their
  compositionality.
\newblock In {\em Advances in neural information processing systems}, pages
  3111--3119, 2013.

\bibitem{mnih2016asynchronous}
V.~Mnih, A.~P. Badia, M.~Mirza, A.~Graves, T.~Lillicrap, T.~Harley, D.~Silver,
  and K.~Kavukcuoglu.
\newblock Asynchronous methods for deep reinforcement learning.
\newblock In {\em International conference on machine learning}, pages
  1928--1937. PMLR, 2016.

\bibitem{mnih2013playing}
V.~Mnih, K.~Kavukcuoglu, D.~Silver, A.~Graves, I.~Antonoglou, D.~Wierstra, and
  M.~Riedmiller.
\newblock Playing atari with deep reinforcement learning.
\newblock {\em arXiv preprint arXiv:1312.5602}, 2013.

\bibitem{pennington2014glove}
J.~Pennington, R.~Socher, and C.~D. Manning.
\newblock Glove: Global vectors for word representation.
\newblock In {\em Proceedings of the 2014 conference on empirical methods in
  natural language processing (EMNLP)}, pages 1532--1543, 2014.

\bibitem{peters2018deep}
M.~E. Peters, M.~Neumann, M.~Iyyer, M.~Gardner, C.~Clark, K.~Lee, and
  L.~Zettlemoyer.
\newblock Deep contextualized word representations.
\newblock {\em arXiv preprint arXiv:1802.05365}, 2018.

\bibitem{platanios2021value}
E.~A. Platanios, A.~Pauls, S.~Roy, Y.~Zhang, A.~Kyte, A.~Guo, S.~Thomson,
  J.~Krishnamurthy, J.~Wolfe, J.~Andreas, et~al.
\newblock Value-agnostic conversational semantic parsing.
\newblock In {\em Proceedings of the 59th Annual Meeting of the Association for
  Computational Linguistics and the 11th International Joint Conference on
  Natural Language Processing (Volume 1: Long Papers)}, pages 3666--3681, 2021.

\bibitem{puig2018virtualhome}
X.~Puig, K.~Ra, M.~Boben, J.~Li, T.~Wang, S.~Fidler, and A.~Torralba.
\newblock Virtualhome: Simulating household activities via programs.
\newblock In {\em Proceedings of the IEEE Conference on Computer Vision and
  Pattern Recognition}, pages 8494--8502, 2018.

\bibitem{puig2020watch}
X.~Puig, T.~Shu, S.~Li, Z.~Wang, J.~B. Tenenbaum, S.~Fidler, and A.~Torralba.
\newblock Watch-and-help: A challenge for social perception and human-ai
  collaboration.
\newblock {\em arXiv preprint arXiv:2010.09890}, 2020.

\bibitem{radford2018improving}
A.~Radford, K.~Narasimhan, T.~Salimans, and I.~Sutskever.
\newblock Improving language understanding by generative pre-training.
\newblock 2018.

\bibitem{radford2019language}
A.~Radford, J.~Wu, R.~Child, D.~Luan, D.~Amodei, I.~Sutskever, et~al.
\newblock Language models are unsupervised multitask learners.
\newblock {\em OpenAI blog}, 1(8):9, 2019.

\bibitem{stable-baselines3}
A.~Raffin, A.~Hill, A.~Gleave, A.~Kanervisto, M.~Ernestus, and N.~Dormann.
\newblock Stable-baselines3: Reliable reinforcement learning implementations.
\newblock {\em Journal of Machine Learning Research}, 22(268):1--8, 2021.

\bibitem{reid2022can}
M.~Reid, Y.~Yamada, and S.~S. Gu.
\newblock Can wikipedia help offline reinforcement learning?
\newblock {\em arXiv preprint arXiv:2201.12122}, 2022.

\bibitem{schulman2017proximal}
J.~Schulman, F.~Wolski, P.~Dhariwal, A.~Radford, and O.~Klimov.
\newblock Proximal policy optimization algorithms.
\newblock {\em arXiv preprint arXiv:1707.06347}, 2017.

\bibitem{sharma22sl3}
P.~Sharma, A.~Torralba, and J.~Andreas.
\newblock Skill induction and planning with latent language.
\newblock In {\em Association for Computational Linguistics}, 2022.

\bibitem{shen2020igibson}
B.~Shen, F.~Xia, C.~Li, R.~Mart{\'\i}n-Mart{\'\i}n, L.~Fan, G.~Wang, S.~Buch,
  C.~D'Arpino, S.~Srivastava, L.~P. Tchapmi, et~al.
\newblock igibson, a simulation environment for interactive tasks in large
  realisticscenes.
\newblock {\em arXiv preprint arXiv:2012.02924}, 2020.

\bibitem{shridhar2020alfred}
M.~Shridhar, J.~Thomason, D.~Gordon, Y.~Bisk, W.~Han, R.~Mottaghi,
  L.~Zettlemoyer, and D.~Fox.
\newblock Alfred: A benchmark for interpreting grounded instructions for
  everyday tasks.
\newblock In {\em Proceedings of the IEEE/CVF conference on computer vision and
  pattern recognition}, pages 10740--10749, 2020.

\bibitem{todorov2012mujoco}
E.~Todorov, T.~Erez, and Y.~Tassa.
\newblock Mujoco: A physics engine for model-based control.
\newblock In {\em 2012 IEEE/RSJ International Conference on Intelligent Robots
  and Systems}, pages 5026--5033. IEEE, 2012.

\bibitem{tsimpoukelli2021multimodal}
M.~Tsimpoukelli, J.~Menick, S.~Cabi, S.~Eslami, O.~Vinyals, and F.~Hill.
\newblock Multimodal few-shot learning with frozen language models.
\newblock {\em arXiv preprint arXiv:2106.13884}, 2021.

\bibitem{vaswani2017attention}
A.~Vaswani, N.~Shazeer, N.~Parmar, J.~Uszkoreit, L.~Jones, A.~N. Gomez,
  L.~Kaiser, and I.~Polosukhin.
\newblock Attention is all you need.
\newblock {\em arXiv preprint arXiv:1706.03762}, 2017.

\bibitem{vig-2019-multiscale}
J.~Vig.
\newblock A multiscale visualization of attention in the transformer model.
\newblock In {\em Proceedings of the 57th Annual Meeting of the Association for
  Computational Linguistics: System Demonstrations}, pages 37--42, Florence,
  Italy, July 2019. Association for Computational Linguistics.

\bibitem{wang2019rat}
B.~Wang, R.~Shin, X.~Liu, O.~Polozov, and M.~Richardson.
\newblock Rat-sql: Relation-aware schema encoding and linking for text-to-sql
  parsers.
\newblock {\em arXiv preprint arXiv:1911.04942}, 2019.

\bibitem{wolf2019huggingface}
T.~Wolf, L.~Debut, V.~Sanh, J.~Chaumond, C.~Delangue, A.~Moi, P.~Cistac,
  T.~Rault, R.~Louf, M.~Funtowicz, et~al.
\newblock Huggingface's transformers: State-of-the-art natural language
  processing.
\newblock {\em arXiv preprint arXiv:1910.03771}, 2019.

\bibitem{yang2019neurips}
Z.~Yang, Z.~Dai, Y.~Yang, J.~Carbonell, R.~R. Salakhutdinov, and Q.~V. Le.
\newblock Xlnet: Generalized autoregressive pretraining for language
  understanding.
\newblock In H.~Wallach, H.~Larochelle, A.~Beygelzimer, F.~d\textquotesingle
  Alch\'{e}-Buc, E.~Fox, and R.~Garnett, editors, {\em Advances in Neural
  Information Processing Systems}, volume~32. Curran Associates, Inc., 2019.

\bibitem{yang2020bert}
Z.~Yang, N.~Garcia, C.~Chu, M.~Otani, Y.~Nakashima, and H.~Takemura.
\newblock Bert representations for video question answering.
\newblock In {\em Proceedings of the IEEE/CVF Winter Conference on Applications
  of Computer Vision}, pages 1556--1565, 2020.

\bibitem{zhu2020incorporating}
J.~Zhu, Y.~Xia, L.~Wu, D.~He, T.~Qin, W.~Zhou, H.~Li, and T.-Y. Liu.
\newblock Incorporating bert into neural machine translation.
\newblock {\em arXiv preprint arXiv:2002.06823}, 2020.

\end{thebibliography}
